\documentclass{article}

\usepackage{arxiv}

\usepackage[utf8]{inputenc} 
\usepackage[T1]{fontenc}    
\usepackage{hyperref}       
\usepackage{url}            
\usepackage{booktabs}       
\usepackage{amsfonts}       
\usepackage{nicefrac}       
\usepackage{microtype}      
\usepackage{lipsum}

\usepackage{amsmath}
\usepackage{setspace}
\usepackage{mathtools}
\usepackage{accents}
\usepackage{amssymb}
\usepackage{enumitem}
\usepackage{hyperref}
\usepackage{enumitem}
\usepackage{longtable}
\usepackage{fixmath}
\usepackage{wrapfig}
\usepackage{color}
\usepackage{algorithm}
\usepackage{algpseudocode}
\usepackage{comment}
\usepackage{subfig}
\usepackage{soul}
\usepackage{lineno}
\usepackage{graphicx,multicol}
\usepackage{url}
\usepackage{hyperref}
\usepackage{multicol,multirow}

\title{Inverse Classification with Limited Budget and Maximum Number of Perturbed Samples}

\author{
  Jaehoon Koo\\
  Department of Industrial Engineering and Management Sciences\\
  Northwestern University\\
  Evanston, IL 60208 \\
  \texttt{jh.koo@u.northwestern.edu} \\
   \And
   Diego Klabjan \\
  Department of Industrial Engineering and Management Sciences\\
  Northwestern University\\
  Evanston, IL 60208 \\
  \texttt{d-klabjan@northwestern.edu} \\
   \And
   Jean Utke \\ 
  Data, Discovery and Decision Science \\
  Allstate Insurance Company\\
  Northbrook, IL 60062 \\
  \texttt{jutke@allstate.com} \\
}

\begin{document}
\maketitle

\begin{abstract}
Most recent machine learning research focuses on developing new classifiers for the sake of improving classification accuracy. With many well-performing state-of-the-art classifiers  available, there is a growing need for understanding interpretability of a classifier necessitated by practical purposes such as to find the best diet recommendation for a diabetes patient. Inverse classification is a post modeling process to find changes in input features of samples to alter the initially predicted class. It is useful in many business applications to determine how to adjust a sample input data such that the classifier predicts it to be in a desired class. In real world applications, a budget on perturbations of samples corresponding to customers or patients is usually considered, and in this setting, the number of successfully perturbed samples is key to increase benefits. In this study, we propose a new framework to solve inverse classification that maximizes the number of perturbed samples subject to a per-feature-budget limits and favorable classification classes of the perturbed samples. We design algorithms to solve this optimization problem based on gradient methods, stochastic processes, Lagrangian relaxations, and the Gumbel trick. In experiments, we find that our algorithms based on stochastic processes exhibit an excellent performance in different budget settings and they scale well.
    
\end{abstract}

\section{Introduction}

Classification is a building block for solving various machine learning tasks such as customer segmentation, sentimental analysis, and image recognition. Numerous state-of-the-art classification models including deep neural networks have been developed to achieve high classification accuracy \cite{Aggarwal2010,Lash2017a}. 
A common post modeling step is to consider changes in features that alter the predicted class, e.g. from the prediction of becoming sick to remaining healthy. 
Given a trained classifier, inverse classification models identify minimal changes of input features of a sample so that the sample is predicted as a desired class that is different from its originally predicted class \cite{Laugel2018}. It is first introduced as a topic of sensitivity analysis \cite{Mannino2000} and then augmented as an interpretability approach \cite{Barbella2009}. Viewing inverse classification as a utility-based data mining problem, Lash et al.~\cite{Lash2017a} argue that it is a subtopic of strategic learning \cite{Boylu2010}. Inverse classification is also related to counterfactual explanation in interpretable machine learning. A counterfactual explanation reveals how a sample should be perturbed to significantly change its original prediction. By crafting counterfactual samples we can interpret how a classifier computes individual predictions \cite{Molnar2019,Wachter2018}. Laugel et al.~\cite{Laugel2018} and Lowd and Meck~\cite{Lowd2005} point out that inverse classification is related to adversarial learning \cite{Tygar2011} that aims to attack a classifier by applying small perturbations to samples to modify their initial predictions. Inverse classification and counterfactual explanation studies focus on interpretability of classification models. Meanwhile, adversarial learning mainly focuses on robustness of associated models. For example, developing a defensive system against adversarial attacks is a study of interest in adversarial learning. Perturbing samples so that they are predicted as a desired label is a common goal in these areas.

In many business applications samples correspond to treatment of customers or patients and the goal is to perturb their treatment in order to be predicted in a more favorable class. We often have limits on feature perturbation amounts, e.g., a limited number of interactions with all of the customers or we only have limited availability of a drug or procedure. Within per-feature-budget limits we want to treat as many customers or patients as possible. This is equivalent to stating that we want to maximize the number of perturbed samples because we turn each one of them into a more favorable class. Past works deal with objectives such as minimizing budget or other continuous loss functions but to capture the number of samples to select yields a discrete function which poses unique challenges addressed herein. In this work we are focusing on the problem of maximizing the number of selected samples subject to per-feature-budget limits and desired reclassification of the perturbed samples. This setting is a direct result of use cases that are encountered by our industry partner. 

A typical setting considered in inverse classification, counterfactual explanations, and adversarial learning is as follows. Let us assume that we have a classifier $f: x \in X \rightarrow y \in Y$ with input $x$ and output $y$. The goal is to generate an adversarial sample $\hat{x}$ that is of the same form as a given sample $x$, and is to be predicted as a desired class that is different from the originally predicted label of $x$. Especially in adversarial learning, there are two types of adversarial examples. A non-targeted adversarial example $\hat{x}$ is generated by adding small perturbation to $x$ so that $\hat{x}$ is classified as any class that is not the original ground truth. A targeted adversarial sample fools a classifier so that it produces a desired label $f(\hat{x}) = \bar{y}$ where $\bar{y}$ is the desired class determined by the adversary. 
The $L_p$ norm of the perturbation between adversarial samples and given samples is usually used as the loss function where $p$ can be $0,1,2,\infty$ \cite{Dong2018}. In some cases, a set of budget constraints for the perturbation is introduced, and subsequently, not all of candidate samples can be successfully perturbed as desired. Herein we focus on spending the budget so that as many samples as possible can be successfully perturbed within the budget. Existing inverse classification and adversarial attack frameworks that minimize cost of the perturbation do not consider and capture this perspective. 

In this study, we develop a new framework that can be applied to inverse classification and counterfactual explanations as well as adversarial learning. We assume to have a budget constraint on perturbations of continuous input features. In order to obtain the maximal number of successfully perturbed samples within a budget, we define an objective function that maximizes the number of samples to be perturbed, which is different from the existing formulations. For this, we introduce a binary variable indicating which sample is to be perturbed. In addition, we include a set of constraints that the probability of the desired class produced by the classifier is higher than all other classes by a tunable margin. This is done to avoid a purely adversarial change in the prediction but instead induce perturbations yielding the actual desired change in the real-world process. We propose Langrangian-based models relying on binary variables. An alternative view considering the selection of samples as a stochastic process with unknown probabilities yields even better algorithms. The resulting model uses chance constraints and the Gumbel trick to make algorithms more efficient. We rely on gradient-based optimization in conjunction with Lagrangian relaxations. 

For evaluation, we use a real-world proprietary dataset from the insurance industry and a public dataset from a health clinic, \cite{mimic2,mimic1}. We compare the performance of our algorithms with respect to the number of selected samples and the budget consumed per selected sample. Algorithms based on stochastic processes significantly outperform those relying on binary variables. We conduct budget and scalability experiments on the real-world data; the relative improvement of the proposed stochastic algorithms over an existing method with a traditional objective function is 7\% and 24\% on average for a variety of budget and possible sample size settings, respectively. In the budget experiments on the public data, the stochastic algorithms outperform the existing model by 19\% on average.

The contributions of this work are as follows.
\begin{enumerate}
 \item We introduce a new framework to solve inverse classification to achieve the maximal number of successfully perturbed samples within a per-feature budget. As far as we know, this framework has not yet been applied to inverse classification in the existing literature.
 \item We design novel algorithms based on gradient methods, stochastic processes, Lagrangian relaxations and the Gumbel trick.
 \item In the computational study, stochastic approaches perform well on different budget scenarios, and they scale. 
\end{enumerate}

The rest of this paper is organized as follows. In Section \ref{sec:lr}, the related work is discussed. Section \ref{sec:model} describes the proposed models, and the algorithms to solve them are presented in Section \ref{sec:comp}. Section \ref{sec:comp} provides the computational study including experimental details and analyses of the experimental results. Conclusions are given in Section \ref{sec:conclusion}.

\section{Related work} \label{sec:lr}

In inverse classification and counterfactual explanation studies the focus is either on unconstrained or constrained problems, or algorithmic mechanisms \cite{Lash2017a, Lash2017b}. A formulation framework is related to feasibility and implementability of perturbed samples, which yield either an unconstrained \cite{Aggarwal2010,Yang2012} or a constrained problem \cite{Barbella2009,Chi2012,Lash2017a,Lash2017b,Mannino2000,Wachter2018}. Since an unconstrained formulation does not consider practical constraints such as a budget, it tends to produce unrealistic perturbations of input features of a sample, e.g. cannot offer a drug if a patient is at home. A constrained formulation provides realistic perturbations, however, it is challenging to solve them. There are three factors to be considered: a) identify changeable features to be perturbed, e.g. an unchangeable feature can be a product purchase history, b) how costly is it to change a feature, and c) limit the amount of perturbations over all samples, i.e., a budget \cite{Lash2017a, Lash2017b}. In \cite{Barbella2009}, only aspect c is considered. Mannino et al. \cite{Mannino2000} consider b), but do not consider a) and c). Lash et al. \cite{Lash2017a} propose a general framework that considers a, b, and c, however, a prediction confidence constraint is not included. With respect to algorithms there are greedy \cite{Aggarwal2010,Chi2012,Lash2017b,Mannino2000,Yang2012} and non-greedy \cite{Barbella2009,Lash2017a} algorithms. Greedy methods are computationally efficient but typically suffer from low solution quality. 
Non-greedy methods tend to focus on more moderate objectives not capturing many aspects so that the obtained adversarial samples are more realistic. 
In \cite{Aggarwal2010,Chi2012,Lash2017b,Mannino2000,Yang2012}, heuristic methods that do not use gradients such as local search, hill climbing, and genetic algorithm are used. In \cite{Lash2017a}, a projected gradient method is adopted and in \cite{Barbella2009}, a non-linear solver package is used to solve a constrained problem. Our work is different from the aforementioned research since none of the existing methods consider maximizing the number of perturbed samples in their formulation. Because of the discrete nature of the counting objective functions these algorithms are not appropriate.

In adversarial learning, recent research mostly focuses on generating adversarial samples to attack deep learning models since its purpose is to study robustness of state-of-the-art classifiers. Most adversarial attacks are targeted against deep networks on visual or audio perception tasks as they allow a striking demonstration of the large effects of minute perturbations on the eventual prediction results in comparison with the robustness of human perception. Szegedy et al. \cite{Szegedy2013} propose the following optimization problem and solve it by using constrained L-BFGS, 
\begin{equation*}
\begin{aligned}
& \underset{\hat{x}}{\text{min}}
\quad c ||x-\hat{x}||_{2}^{2} +J(\hat{x})\\ 
& \text{s.t.}
\quad \hat{x} \in [0,1] \\
\end{aligned}
\end{equation*}
where $J$ is a loss function of classification toward a desired label such as cross-entropy and $c$ is hyper parameter. Goodfellow et al. \cite{Goodfellow2015} propose a method called Fast Gradient Sign method (FGSM) using the sign of gradients, where $L_{\infty}$ is used as a distance metric for perturbation. It is not guaranteed to produce optimal solutions, but it is quick to obtain close adversarial examples. Given an input image $x$ FGSM performs \[ \hat{x} = x - \epsilon \: \text{sign} (\nabla J (x)) \]
where $\epsilon$ is selected to be small. Kurakin et al. \cite{Kurakin2017} propose an improved FGSM, iterative gradient sign method (I-FGSM), by taking multiple smaller steps $\alpha$ in the direction of gradient sign rather than one step and its output is clipped by $\epsilon$. It produces superior results to FGSM by updating $\hat{x}$ on each iteration $t$ as
\[ \hat{x}_{t} = \hat{x}_{t-1} - \text{clip}_{\epsilon} (\alpha \: \text{sign} (\nabla J (\hat{x}_{t-1}))) \]
where $\hat{x}_0 = x$.
Recently, Papernot et al. \cite{Papernot2016} propose a greedy algorithm to generate adversarial examples using gradients to compute a saliency map, called Jacobian-based Saliency Map Attack (JSMA). 
These three algorithms do not consider a budget constraint that is critical and practical in inverse classification even though they have a bound at a pixel level. In addition, our framework is designed to achieve the maximal number of successfully perturbed samples within a budget. These algorithms cannot be modified in a meaningful way to tackle our setting. 

\section{Proposed models}  \label{sec:model}

In this section, we present our constrained optimization problem to solve inverse classification. The formulation is designed to generate the maximal number of adversarial examples that are classified as a desired class. In addition, we include a set of budget constraints on perturbations of input features. We first introduce notation and the baseline model. Next, we present variations of the baseline model - chance constraint models that assume decision variables follow an unknown probability distribution.

We denote a given sample by $\mathbf{x} \in \mathbb{R}^p$ and a perturbed sample by $\hat{\mathbf{x}} \in \mathbb{R}^p$. We assume that all features are continuous. Let $f(\mathbf{x}) : \mathbb{R}^p \mapsto [0,1]^k$ be a function associated with a classification model that computes a score of $\mathbf{x}$ being in a class such as the probability of $k$ classes. We optimize over samples $\hat{\mathbf{x}}_j$ given a new desired label vector $\bar{\mathbf{y}}_j \in [0,1]^{k}$ with $j = 1, \ldots, |{\cal S}|$ and ${\cal S} = \{\mathbf{x}_1, \mathbf{x}_2, \ldots, \mathbf{x}_{|\cal S|}\}$. We denote a perturbed input feature matrix by $\hat{\mathbf{X}} = (\hat{\mathbf{x}}_1, \hat{\mathbf{x}}_2, \ldots, \hat{\mathbf{x}}_{|\cal S|} )$. Furthermore, we introduce binary variables, $\mathbf{z} \in \{0,1\}^{|{\cal S}|} $, to decide which sample is to be perturbed. 
We maximize the number of these binary variables that have value 1. We introduce general budget constraints on perturbations of input features. In addition, we have a per-sample constraint, called the prediction confidence constraint, to capture a margin for prediction reflecting the uncertainty in the score function. Formally, the max samples model (MS) is formulated as
\begin{equation} \label{model:MS}
\begin{aligned}
& \underset{\begin{subarray}{c}
  \mathbf{z},\hat{\mathbf{X}}
  \end{subarray}}{\text{max}}
\quad \sum_{j=1}^{|{\cal S}|} z_j\\
& \text{s.t.} \quad g_i(\mathbf{z},\hat{\mathbf{x}}) \leq 0, i=1, \ldots, p, \\
& \quad \quad \,  h_j(\mathbf{z},\hat{\mathbf{x}}) \leq 0, j=1, \ldots, |\cal S|
\end{aligned}
\end{equation} 
where each $g_i$ and $h_j$ is a nonlinear function associated with the budget and prediction confidence constraint, respectively. A typical budget constraint for feature $i$ on perturbation of input features is  
\begin{equation} \label{model:budget}
\begin{aligned}
g_i(\mathbf{z},\hat{\mathbf{x}}) = \sum_{j=1}^{|S|} z_j\:||\hat{x}_{ij} - x_{ij} ||_2^2 - B_i
\end{aligned}
\end{equation}
where $B_i \in \mathbb{R}_{+}$ is a given budget for feature $i$. Prediction confidence constraints are explicitly expressed as 
\begin{equation} \label{model:conf0}
\begin{aligned}
f(\hat{\mathbf{x}}_j)_u + \delta \leq f(\hat{\mathbf{x}}_j)_{\tilde{y}_j} \mbox{ with }\; j=1, \ldots, |{\cal S}|, u=1, \ldots, k, u \neq \tilde{y}_j
\end{aligned}
\end{equation}
where $\tilde{y}_j \in \underset{u}{\text{argmax}} \: [\bar{\mathbf{y}}_j]_u $ is a desired class of $\mathbf{x}_j$ and $\delta >0$ is a given margin. In MS, we rewrite (\ref{model:conf0}) by multiplying it with binary variables so that we consider the constraints only on selected samples as follows: 
\begin{equation} \label{model:conf}
\begin{aligned}
h_j(\mathbf{z},\hat{\mathbf{x}}) = z_j \cdot \bar{h}_j(\hat{\mathbf{x}})= z_j\: \text{max} \{0,\underset{ \substack{u=1, \ldots, k \\ u \neq \tilde{y}_j}}{\text{max}} \: f(\hat{\mathbf{x}}_{j})_u - f(\hat{\mathbf{x}}_{j})_{\tilde{y}_j} + \delta \}, j = 1, \ldots, |{\cal S}|.
\end{aligned}
\end{equation} 
MS is challenging to solve due to the presence of constraints and binary variables $\mathbf{z}$. To address the latter, we assume that the binary variables have a probability distribution which is an approximation. We impose that the variables follow either Bernoulli or Categorical distributions considering dependency among samples. First, we present the Bernoulli case where no relationship among perturbed samples is assumed. 

Let $z_j \sim \text{Bernoulli}(\pi_j), j = 1, \ldots, |{\cal S}|$, i.e., $z_j = 1$ with probability $\pi_j$. Let $0 \leq \Pi = (\pi_1, \pi_2, \ldots, \pi_{ |\cal S|}) \leq 1$.
Transforming MS, we propose a Bernoulli chance max samples model (BCMS) as 
\begin{equation} \label{model:BCMS}
\begin{aligned}
& \underset{\Pi, \hat{\mathbf{X}} }{\text{max}}
\quad  \mathbb{E}_{\mathbf{z}} \bigg[  \sum_{j = 1}^{|{\cal S}|} z_j \bigg] \\  
& \text{s.t.} \quad \text{Pr} \big(g_i(\mathbf{z},\hat{\mathbf{x}}) \leq 0 \big) \geq  1 - \epsilon, i=1, \ldots, p, \\
& \quad \quad \, \text{Pr} \big(h_j(\mathbf{z},\hat{\mathbf{x}}) \leq 0 \big) \geq  1 - \epsilon, j=1, \ldots, |{\cal S}|, 
\end{aligned}
\end{equation}
where $\epsilon$ is a parameter. We can explicitly rewrite BCMS as
\begin{equation} \label{model:BCMS1}
\begin{aligned}
& \underset{\Pi, \hat{\mathbf{X}} }{\text{max}}
\quad \sum_{j=1}^{|{\cal S}|} \pi_j \\
& \text{s.t.}
\quad \text{Pr} \big( \sum_{j=1}^{|S|} z_j\:||\hat{x}_{ij} - x_{ij} ||_2^2 - B_i \leq 0 \big) \geq  1 - \epsilon, i = 1, \ldots, p\\
& \quad  \quad \: \pi_j\: \bar{h}_j (\hat{\mathbf{x}}_{j}) \leq 0, j = 1, \ldots, |{\cal S}|.
\end{aligned}
\end{equation}
The chance max samples model with the Categorical distribution (CCMS) considers dependency among samples to be perturbed. To this end, let $0 \leq \Pi = (\pi_1, \pi_2, \ldots, \pi_{ |\cal S|})$ with $\sum_{j=1}^{|\cal S|} \: \pi_j = 1$ and $K$ an integer parameter. CCMS has the same formulation as BCMS, but we use the following binary variables to determine which sample to perturb:
\begin{equation} \label{model:Cat} 
\begin{aligned}
\bar{z}_{\xi} \in \mathbb{R}^{|{\cal S}|} \sim \text{Cat} (\Pi), \xi = 1, \ldots, K \text{ and } \mathbf{z} \in \mathbb{R}^{|{\cal S}|}, \mathbf{z} = \text{min} (\mathbf{1}, \sum_{\xi=1}^{K} \bar{z}_{\xi}).
\end{aligned}
\end{equation}
We finish with a benchmark model based on the existing framework \cite{Molnar2019,Szegedy2013} of generating adversarial samples that is designed to solve inverse classification with a minimal cost of perturbation. This model optimizes over samples in $\hat{\mathbf{x}}_j \in \cal S$ to minimize the loss function $l_j(\hat{\mathbf{x}}) = KL \: (\bar{\mathbf{y}}_j|| f(\hat{\mathbf{x}}_j)) + a \: ||\hat{\mathbf{x}}_j - \mathbf{x}_j ||_2^2 $, $a \in [0,\infty)$ where $KL$ denotes the Kullback-Leibler divergence. The model (KL) reads 
\begin{equation} \label{model:KL}
\begin{aligned}
& \underset{\hat{\mathbf{X}}}{\text{min}}
\quad \sum_{j = 1}^{|{\cal S}|} l_j \:(\hat{\mathbf{x}}) \\
& \text{s.t.} \quad g_i(\hat{\mathbf{x}}) \leq 0, i=1, \ldots, p, \\
& \quad \quad \: \bar{h}_{j}(\hat{\mathbf{x}}) \leq 0, j=1, \ldots, |{\cal S}|
\end{aligned}
\end{equation}
where each $g$ and $h$ is a nonlinear function associated with budget (\ref{model:budget}) and prediction confidence constraints (\ref{model:conf0}) without the binary variable $\mathbf{z}$. This model has a smaller number of decision variables than our models since it does not include binary variables or any possible related variables; however, it does not explicitly count the number of perturbed samples and thus it has a different objective. KL does not necessarily achieve the maximal number of successfully perturbed samples.

\section{Algorithms} \label{sec:alg}

In this section, we describe algorithms based on Lagrangian and subgradient methods. We first reformulate the problem as an unconstrained problem by using Lagrangian multipliers. We develop algorithms to solve Lagrangian functions based on the projected subgradient method. Projection is used to keep Lagrangian multipliers positive during updates or to maintain probability requirements. For chance max samples models, we apply the Gumbel trick \cite{Maddison2014} to use approximate gradients when updating the binary variables. 

\paragraph{Algorithm for max samples model}

We first define the Lagrangian function for MS (\ref{model:MS}) as 
\begin{equation*}
\begin{aligned}
& L(\mathbf{z},\hat{\mathbf{X}},\lambda, \mu) = \sum_{j = 1}^{|{\cal S}|} z_j - \sum_{i=1}^{p} \lambda_i g_i(\mathbf{z},\hat{\mathbf{x}}) - \sum_{j=1}^{|\cal S|} \mu_j h_j(\mathbf{z},\hat{\mathbf{x}})  \\ 
& \quad \quad = \sum_{j = 1}^{|{\cal S}|} z_j - \; \sum_{i=1}^{p} \lambda_i \big( \sum_{j=1}^{|S|} z_j\:||\hat{x}_{ij} - x_{ij} ||_2^2 - B_i \big) - \sum_{j = 1}^{|{\cal S}|} \mu_{j} z_j \bar{h}_{j} (\hat{\mathbf{x}}_j) \\
& \quad \quad = \sum_{j = 1}^{|{\cal S}|} c_j z_j  + \sum_{i=1}^{p} \lambda_i B_i
\end{aligned}
\end{equation*}
where
\begin{equation} \label{eqn:cj}
c_j = 1 - \sum_{i=1}^{p} \lambda_i\:||\hat{x}_{ij} - x_{ij} ||_2^2 - \mu_{j} \bar{h}_{j} (\hat{\mathbf{x}}_j)
\end{equation}
and $\lambda$ and $\mu$ are Lagrangian multipliers. 
We propose Algorithm \ref{alg:MS} to solve $\underset{\lambda, \mu}{\text{min}} \: \underset{\mathbf{z},\hat{\mathbf{X}}}{\text{max}} \: L(\mathbf{z}, \hat{\mathbf{X}},\lambda, \mu)$. The algorithm consists of two main loops to solve the $ \text{min} \: \text{max}$ problem; the inner loop updates input features and binary variables $\mathbf{z}$ to maximize $L$, and  the outer loop updates Lagrangian multipliers to minimize $L$. In the algorithm, we initialize all $z$ with one as we aim to achieve as many successfully perturbed samples as possible. Meanwhile, we add a line to break the inner loop when all entries of $\mathbf{z}$ are zero, which is the case of no updates on variables.
\begin{algorithm} 
\caption{MS}
    \label{alg:MS}
    \begin{algorithmic}[1]
    \State Initialize $\lambda, \mu, \gamma, \eta$. 
        \While{until convergence}
            \State {$z_j \leftarrow 1, j=1,\ldots, |{\cal S}|$}
            \While{until convergence}
                   \State {Break \textbf{if} $\mathbf{z} = 0$ }
                   \State {$\hat{\mathbf{X}}^{*} \leftarrow \underset{\hat{\mathbf{X}}}{\text{argmax}} \; L(\mathbf{z},\hat{\mathbf{X}}, \lambda, \mu) \mbox{ and } c_j \mbox { as in (\ref{eqn:cj})}$}
                   \For{$j=1,\ldots,|{\cal S}|$}
                        \If{$c_j \geq 0 $}
                           \State {$z_j \leftarrow 1$}
                        \Else
                           \State {$z_j \leftarrow 0$}
                        \EndIf
                   \EndFor
            \EndWhile
            \State {$\lambda_{i} \leftarrow \big(\lambda_{i} - \gamma \: \nabla_{\lambda_{i}} L \big)^{+}, i = 1,\ldots, p$}
            \State {$\mu_{j} \leftarrow \big( \mu_{j} - \eta \: \nabla_{\mu_{j}} L \big)^{+}, j = 1,\ldots, |{\cal S}|$}
        \EndWhile
    \end{algorithmic}
\end{algorithm}
Note that $\nabla_{\lambda_{i}} L = - \sum_{j=1}^{|S|} z_j\:||\hat{x}_{ij} - x_{ij} ||_2^2 + B_i$, and $\nabla_{\mu_{j}} L = - z_j \bar{h}_{j} (\hat{\mathbf{x}}_j) $. In addition, lines 7-13 in Algorithm \ref{alg:MS} are derived by solving 
\begin{equation*}
\begin{aligned}
& \underset{ \mathbf{z} }{\text{max}}
\quad \sum_{j=1}^{|{\cal S}|} c_j z_j
\end{aligned}
\end{equation*}
where $c_j$ is assumed to be constant in this part of the algorithm. 

\paragraph{Algorithm for Bernoulli and Categorical chance max samples model}

We define the Lagrangian function for BCMS (\ref{model:BCMS1}) as
\begin{equation} 
\label{lag:BCMS} 
\begin{aligned}
& L(\Pi, \hat{\mathbf{X}},\lambda, \mu) = \sum_{j = 1}^{|{\cal S}|} \pi_j \big( 1 - \mu_{j} \bar{h}_{j} (\hat{\mathbf{x}}_j) \big) + \sum_{i=1}^{p} \lambda_i \bigg[ \text{Pr} \big( \sum_{j=1}^{|S|} z_j\:||\hat{x}_{ij} - x_{ij} ||_2^2 - B_i \leq 0 \big) - (1- \epsilon) \bigg]. \\
\end{aligned}
\end{equation}
We need to solve $\underset{\lambda, \mu}{\text{min}} \: \underset{\Pi,\hat{\mathbf{X}}}{\text{max}} \: L$ where we have to compute gradients of $\mathbb{E}_{\mathbf{z}\sim \text{Ber}(\Pi)}$ with respect to $\Pi$. To relieve the burden of computing exact gradients, we apply the Gumbel trick \cite{Maddison2014} to use their approximation. Let the exact probability be $\text{Pr}_i = \mathbb{E}_{\mathbf{z}} {\cal X} (\sum_{j=1}^{|S|} z_j\:||\hat{x}_{ij} - x_{ij} ||_2^2 - B_i \leq 0)$ where ${\cal X}$ is the indicator function. We first approximate ${\cal {X}} (\mathbf{x}) \approx ({1 + \text{exp}({- \kappa  
\frac{ \mathbf{x} -\tau}{1-\tau}} )})^{-1}$ where $ \kappa$ and $\tau$ are hyperparameters. We have 
$\text{Pr}_i \approx \mathbb{E}_{\mathbf{z}} ({1 + \text{exp}({- \kappa  \frac{\mathbf{x}-\tau}{1-\tau}}))^{-1}}$ with $\mathbf{x}= \sum_{j=1}^{|S|} z_j\:||\hat{x}_{ij} - x_{ij} ||_2^2 - B_i$. Let us consider first the Bernoulli case. Applying the Gumbel trick, the expectation term is approximately computed as
\begin{equation} 
\label{eq:gumbel_app} 
\begin{aligned}
\text{Pr}_i \approx \frac{1}{N} \sum_{n=1}^{N} \text{P}_n^i = \frac{1}{N} \sum_{n=1}^{N}  (1 + \text{exp}({- \kappa  \frac{\mathbf{x}_n^{i}-\tau}{1-\tau}} ))^{-1}
\end{aligned}
\end{equation}
where 
\begin{equation*}
\mathbf{x}_n^{i} = \sum_{j=1}^{|S|} v_{nj} \:||\hat{x}_{ij} - x_{ij} ||_2^2 - B_i  \mbox{ and } v_{nj} = \bar{z}_j (\pi_j, g_1^{n}, g_2^{n}) =  \frac{\text{exp}((\text{log}\pi_j +g_1^{n})/\omega)}{ \text{exp}((\text{log}\pi_j +g_1^{n} )/\omega) + \text{exp}((\text{log}(1-\pi_j) +g_2^{n} )/\omega)}
\end{equation*}
and $g_1^{n} \sim {\cal G}, g_2^{n} \sim {\cal G}$. Here the Gumbel distribution is denoted by ${\cal G}$ and $\omega$ and $N$ are hyperparameters. Values $\bar{z}_j$ approximate $z_j$. We can rewrite (\ref{lag:BCMS}) as
\begin{equation} 
\label{lag:BCMS2} 
\begin{aligned}
& L \approx \sum_{j = 1}^{|{\cal S}|} \pi_j \big( 1 - \mu_{j} \bar{h}_{j} (\hat{\mathbf{x}}_j) \big) + \frac{1}{N} \sum_{n=1}^{N} \sum_{i=1}^{p} \lambda_i \text{P}_n^i - (1 - \epsilon) \sum_{i=1}^{p} \lambda_i.
\end{aligned}
\end{equation}
We propose Algorithm \ref{alg:BCMS} to solve $\underset{\lambda, \mu}{\text{min}} \: \underset{\Pi,\hat{\mathbf{X}}}{\text{max}} \: L$. This algorithm has two main loops to maximize $L$ with respect to $\Pi$ and $\hat{\mathbf{X}}$, and to minimize $L$ with respect to the Lagrangian multipliers. A part of generating Gumbel's samples is added to the inner loop so that approximated gradients are used to update variables. In the algorithm we use 
\begin{equation}
    \label{eqn:ddotL}
\ddot{L} = \sum_{j = 1}^{|{\cal S}|} \pi_j \big( 1 - \mu_{j} \bar{h}_{j} (\hat{\mathbf{x}}_j) \big)
\end{equation}
\begin{algorithm} 
\caption{BCMS}
    \label{alg:BCMS}
    \begin{algorithmic}[1]
    \State Initialize $\Pi, \lambda, \mu, \alpha, \beta, \gamma, \eta$. 
        \While{until convergence}
            \While{until convergence}
                   \For{$n=1,\ldots,N$}
                        \For{$j = 1,\ldots, |{\cal S}|$} 
                        \State {$g_1^{n} \sim {\cal G}_j, g_2^{n} \sim {\cal G}_j $}
                        \State {$ v_{nj} \leftarrow \bar{z}_j(\pi_j,g_1^{n}, g_2^{n}) $}
                        \EndFor
                    \State {$ \dot{L}_n \leftarrow \sum_{i=1}^{p} \lambda_i \text{P}_n^i$ with $\text{P}_n^i$ as in (\ref{eq:gumbel_app})}    
                   \EndFor
            \State {$\nabla_{\Pi} L \leftarrow \frac{1}{N} \sum_{n=1}^{N} \nabla_{\Pi} \dot{L}_n + \nabla_{\Pi} \ddot{L}$ with $\ddot{L}$ as in (\ref{eqn:ddotL})}
            \State {$\Pi \leftarrow \text{min} \{1,\big(\Pi + \alpha \: \nabla_{\Pi} L \big)^{+}\}$}
            \State {$\hat{\mathbf{X}} \leftarrow \hat{\mathbf{X}} + \beta \:  \nabla_{\hat{\mathbf{X}}} L$}
            \EndWhile
            \State {$\lambda_{i} \leftarrow \big(\lambda_{i} - \gamma \: \nabla_{\lambda_{i}} L \big)^{+}, i = 1,\ldots, p$}
            \State {$\mu_{j} \leftarrow \big( \mu_{j} - \eta \: \nabla_{\mu_{j}} L \big)^{+}, j = 1,\ldots, |{\cal S}|$} 
        \EndWhile
    \end{algorithmic} 
\end{algorithm}
For CCMS, we define the Lagrangian function based on (\ref{model:BCMS}), which is written as
\begin{equation*} 
\label{lag:CCMS} 
\begin{aligned}
& L (\Pi, \hat{\mathbf{X}},\Lambda, M) = \mathbb{E}_{\mathbf{z}} \bigg[  \sum_{j = 1}^{|{\cal S}|} z_j \big( 1 - \mu_{j} \bar{h}_{j} (\hat{\mathbf{x}}_j) \big) + \sum_{i=1}^{p} \lambda_i \bigg[\text{Pr} \big( \sum_{j=1}^{|S|} z_j\:||\hat{x}_{ij} - x_{ij} ||_2^2 - B_i \leq 0 \big) - (1- \epsilon) \bigg] \bigg]
\end{aligned}
\end{equation*}
where $z_j$ is $j^{\text{th}}$ element of $\mathbf{z}$ as defined in (\ref{model:Cat}).
Based on the Gumbel's approach we approximate $\mathbf{z}$ by 
\[[\bar{z}_{\xi}]_j = \frac{\text{exp}((\text{log}\pi_j +g_{j\xi})/\omega)}{ \sum_{k=1}^{|{\cal S}|} \text{exp}((\text{log}\pi_k +g_{k\xi} )/\omega) } \text{ where } g_{1\xi}, \ldots, g_{|{\cal S}|\xi} \sim {\cal G}, \xi = 1, \ldots, K. \] 
The approximate Lagrangian function for CCMS reads 
\begin{equation}
\label{lag:CCMS2}
\begin{aligned}
& L \approx \mathbb{E}_{\mathbf{G} \sim {\cal G}} \bigg[ \sum_{j = 1}^{|{\cal S}|} \bar{z}_j (\Pi, \mathbf{G}) \big( 1 - \mu_{j} \bar{h}_{j} (\hat{\mathbf{x}}_j) \big) + \sum_{i=1}^{p} \lambda_i \bigg[ \text{Pr} \big( \sum_{j=1}^{|{\cal S}|} \bar{z}_j (\Pi,\mathbf{G}) \:||\hat{x}_{ij} - x_{ij} ||_2^2 - B_i \leq 0 \big) - (1- \epsilon) \bigg] \bigg] \\
& \approx \frac{1}{N} \sum_{n=1}^{N} \sum_{j = 1}^{|{\cal S}|} v_{nj} \big( 1 - \mu_{j} \bar{h}_{j} \big) + \frac{1}{N} \sum_{n=1}^{N} \sum_i^p \lambda_i \text{P}_n^i - (1 - \epsilon) \sum_{i=1}^{p} \lambda_i 
 = \frac{1}{N} \sum_{n=1}^{N} \Tilde{L}_n + \frac{1}{N} \sum_{n=1}^{N} \dot{L}_n - (1 - \epsilon) \sum_{i=1}^{p} \lambda_i
\end{aligned}
\end{equation}
where $\mathbf{G} \in \mathbb{R}^{|{\cal S}|\times K}$ is a Gumbel matrix, and $\text{P}_n^i$ is the same as in (\ref{eq:gumbel_app}) except that \[ v_{nj} = \bar{z}_j(\Pi,\mathbf{G}^n) = \text{min} (1, \sum_{{\xi}=1}^{K} \frac{\text{exp}((\text{log}\pi_j +g_{j\xi}^{n})/\omega)}{ \sum_{k=1}^{|{\cal S}|} \text{exp}((\text{log}\pi_k +g_{k\xi}^{n} )/\omega) }). \]
We propose Algorithm \ref{alg:CCMS} to solve $\underset{\lambda, \mu}{\text{min}} \: \underset{\Pi,\hat{\mathbf{x}}}{\text{max}} \: L $, which has the same structure as Algorithm \ref{alg:BCMS} for BCMS. The part of simulating Gumbel's samples is edited for the Categorical distribution (lines 4-13).

\begin{algorithm} 
\caption{CCMS}
    \label{alg:CCMS}
    \begin{algorithmic}[1]
    \State Initialize $\Pi, \lambda, \mu, \alpha, \beta, \gamma, \eta$. 
        \While{until convergence}
            \While{until convergence}
                    \For{$n=1,\ldots,N$}
                        \For{$j = 1,\ldots, |{\cal S}|$} 
                            \For {$\xi = 1,\ldots, K$}
                                 \State {$[\mathbf{G}^n]_{j\xi} \leftarrow g_{j\xi}^{n} \sim {\cal G} $}
                            \EndFor
                        \EndFor
                    \State {$ v_{nj} \leftarrow \bar{z}_j(\Pi,\mathbf{G}^n), j = 1, \ldots, |{\cal S}|  $}
                    \State {$\Tilde{L}_n \leftarrow \sum_{j = 1}^{|{\cal S}|} v_{nj} \big( 1 - \mu_{j} \bar{h}_{j} \big)$}
                    \State {$\dot{L}_n \leftarrow \sum_{i=1}^{p} \lambda_i \text{P}_n^i $}
                   \EndFor
            \State {$\nabla_{\Pi} L \leftarrow \frac{1}{N} \sum_{n=1}^{N} (\nabla_{\Pi} \Tilde{L}_n + \nabla_{\Pi} \dot{L}_n)  $}
            \State {$\Pi \leftarrow \big(\Pi + \alpha \: \nabla_{\Pi} L \big)^{+}$}
            \State {$\pi_j \leftarrow \frac{\pi_j}{\sum_i \pi_i}, j = 1,\ldots, |{\cal S}|$}
            \State {$\hat{\mathbf{X}} \leftarrow \hat{\mathbf{X}} + \beta \: \nabla_{\hat{\mathbf{X}}} L$}
            \EndWhile
            \State {$\lambda_{i} \leftarrow \big(\lambda_{i} - \gamma \: \nabla_{\lambda_{i}} L \big)^{+}, i = 1,\ldots, p$}
            \State {$\mu_{j} \leftarrow \big( \mu_{j} - \eta \: \nabla_{\mu_{j}} L \big)^{+}, j = 1,\ldots, |{\cal S}|$}
        \EndWhile
    \end{algorithmic}
\end{algorithm}

\paragraph{Algorithm for KL}

The Lagrangian function for KL (\ref{model:KL}) reads
\begin{equation} \label{lag:KL}
\begin{aligned}
& L(\hat{\mathbf{X}},\lambda, \mu) = \sum_{j = 1}^{|{\cal S}|} l_j \:(\hat{\mathbf{x}}_j,\bar{\mathbf{y}}_j) + \; \sum_{i=1}^{p} \lambda_i \; g_i(\hat{\mathbf{x}}) +  \sum_{j=1}^{|\cal S|} \mu_{j} \; \bar{h}_{j}(\hat{\mathbf{x}})
\end{aligned}
\end{equation}
where $\lambda$ and $\mu$ are Lagrangian multipliers.
Algorithm \ref{alg:KL} is designed to solve $\underset{\lambda, \mu}{\text{max}} \: \underset{\hat{\mathbf{X}}}{\text{min}} \: L$. Similar to Algorithm \ref{alg:MS}, it consists of two loops; the inner loop updates input features to minimize $L$, and the outer loop updates Lagrangian multipliers to maximize $L$. 
\begin{algorithm} 
\caption{KL}
    \label{alg:KL}
    \begin{algorithmic}[1]
    \State Initialize $\lambda, \mu, \gamma, \eta$ 
        \While{until convergence}
            \State {$\hat{\mathbf{X}} \leftarrow \underset{\hat{\mathbf{X}}}{\text{argmin}} \; L(\hat{\mathbf{X}}, \lambda, \mu)$}
            \State {$\lambda_{i} \leftarrow \big(\lambda_{i} + \gamma \: \nabla_{\lambda_{i}} L \big)^{+}, i = 1,\ldots, p$}
            \State {$\mu_{j} \leftarrow \big(\mu_{j} + \eta \: \nabla_{\mu_{j}} L \big)^{+}, j = 1,\ldots, |{\cal S}|$}
        \EndWhile
    \end{algorithmic}
\end{algorithm}

\paragraph{Obtaining the final solution} \label{sec:evaluation}

Since Lagrangian methods do not necessarily guarantee feasibility of solutions $\hat{\mathbf{X}}$ with respect to budget and prediction confidence, we conduct the following steps to obtain the final solution set.
\begin{enumerate}
              \item Run one of the Algorithms 1-4 to obtain a `good' but possibly infeasible solution $\hat{\mathbf{X}}$. 
              \item Find a subset $\Hat{{\cal S}} \subseteq {\cal S}$ of samples satisfying prediction confidence constraints, i.e., all of the samples in $\Hat{{\cal S}}$ are classified as desired. 
              \item Solve the following problem over $\Hat{{\cal S}}$ to find the final set $\Tilde{{\cal S}}$ of feasible samples:
\begin{equation}
\begin{aligned} \label{model:mip}
& \underset{z_1, \ldots, z_{|\Hat{{\cal S}}|}}{\text{max}}
& &  \sum_{s=1}^{|\Hat{{\cal S}}|} z_s\\
& \text{s.t.}
& & \sum_{s=1}^{|\Hat{{\cal S}}|} a_{is}\: z_s \leq B_i, i = 1, \ldots, p\\
& & & z_s \in \{0,1\}, s=1,\dots,|\Hat{{\cal S}}|,
\end{aligned}
\end{equation}
where $a_{is} = ||\hat{x}_{is}-x_{is}||_2^2$. 
          \end{enumerate}

\paragraph{Sequences}

Let us consider the case when samples are sequences of varying length of feature vectors of the same length; e.g., $f$ corresponds to an LSTM or transformer. In this case, $\hat{\mathbf{x}}$ is not well defined, i.e. it is not a matrix and thus (\ref{model:MS}) is ill-posed. If we consider only sequences of the same length, then (\ref{model:MS}) is well defined. If we have $R$ different lengths, then we can form $R$ different disjoint subsets ${\cal S}^r$ of samples and define (\ref{model:MS}) for each one subset. The link between all subsets becomes a joint per-feature budget. This budget needs to be allocated to the $R$ problems. Herein we use a simple strategy of allocating $\frac{|{\cal S}^r|}{|\cal S| } \: B_i$ for each feature $i$ (note that $|\cal S|$ = $\sum_{r=1}^{R} |{\cal S}^r|$).

\section{Computational study} \label{sec:comp}

In this section, we conduct a computational study on two datasets: a proprietary dataset and a public dataset. We experiment with different budget scenarios and we assess scalability with respect to the number of samples. Model implementations for all the experiments are done in Python using Tesla V100 GPU and Intel Xeon CPU E5-2697 v4 @ 2.30Hz for the real-world dataset, and Titan XP 1080 GPU and Intel Xeon Silver 4112 CPU @ 2.60GHz for the public dataset. 

We use the following hyperparameter values: $\delta = 0.1$, $\kappa = 2$, $\tau \in \{30, 50\}$, $\omega=1$, and $N=100$. The learning rates $\alpha$ and $\beta$ affecting $\hat{\mathbf{x}}$ and $\mathbf{z}$ are selected as one of $\{0.01, 0.05, 0.1\}$. We use a decaying learning rates $\gamma$ and $\eta$, affecting the Lagrangian multipliers $\lambda$ and $\mu$, initially set as one in $\{0.5, 1\}$. The stopping criterion is set to be the maximum number of iterations (variable updates). For the outer loop it is set to be 10 for MS, BCMS, and KL and 20 for CCMS, and for the inner loop it is set to be 10,000, 100, 100, and 5,000 for MS, BCMS, CCMS and KL, respectively. The initial Lagrangian multipliers are selected as one from $\{1,10\}$ but adding white Gaussian noise. 
 

\subsection{Real-world data} 
We conduct experiments on a real-world proprietary dataset that contains sequential input features for 5 classes, which is introduced in \cite{Stec2018}. The data has 169 features and sequences are of size from 1 to 150 and thus the joint per-feature budget needs to be employed. The classification model Sparse Time LSTM from \cite{Stec2018} is used as $f$. The accuracy of the model on approximately 700,000 training samples is around 70\%. 

\subsubsection{Budget experiments} \label{sec:as_bud}

We perturb 300 samples selected from the test set which are grouped into 5 different groups by input sequence length. Thus, we have $|{\cal S}| = 300$, and $R = 5$. The 300 samples are correctly predicted by the trained classifier into a "negative" class - four of the five classes -  (e.g. have disease) and thus the perturbed samples should fall into the positive class - the remaining class - (e.g. does not have the disease). We perturb 19 features since the remaining features cannot be altered in practice. To decide the sizes of the budgets, we first run Algorithm \ref{alg:KL} with unlimited budgets to measure how much budget is needed for successful perturbation. Then, as well as based on practical considerations from subject matter experts and data stockholders, we determine small, middle, and large sizes of budgets by the amounts that are proportional to the total budget consumption with the unlimited budgets.

\begin{figure} 
    \centering
  \subfloat[Size of $\Tilde{\cal S}$]{%
      \includegraphics[width=0.45\linewidth]{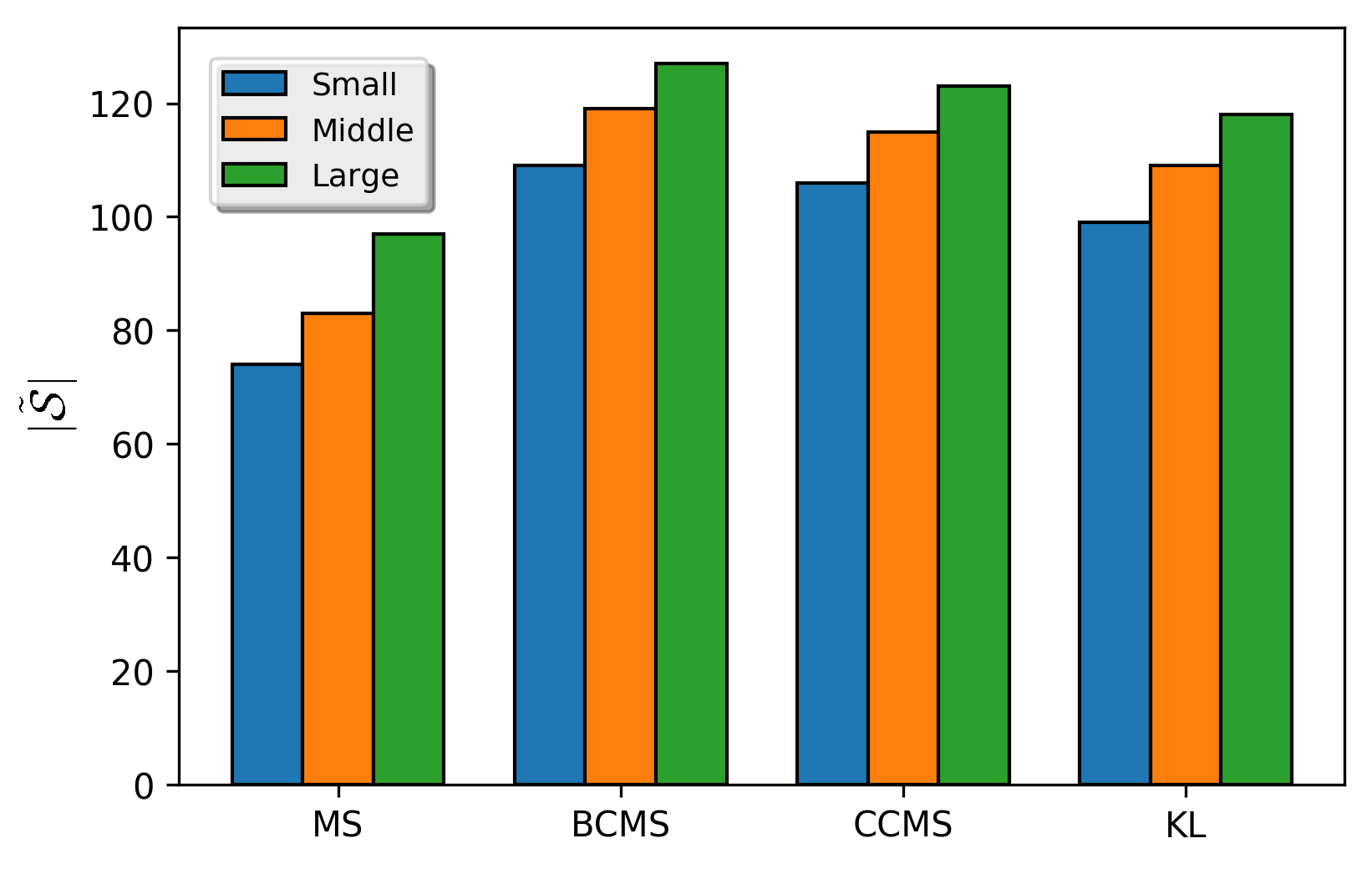}} \hspace{0.6cm}
  \subfloat[Consumption per sample average]{%
      \includegraphics[width=0.45\linewidth]{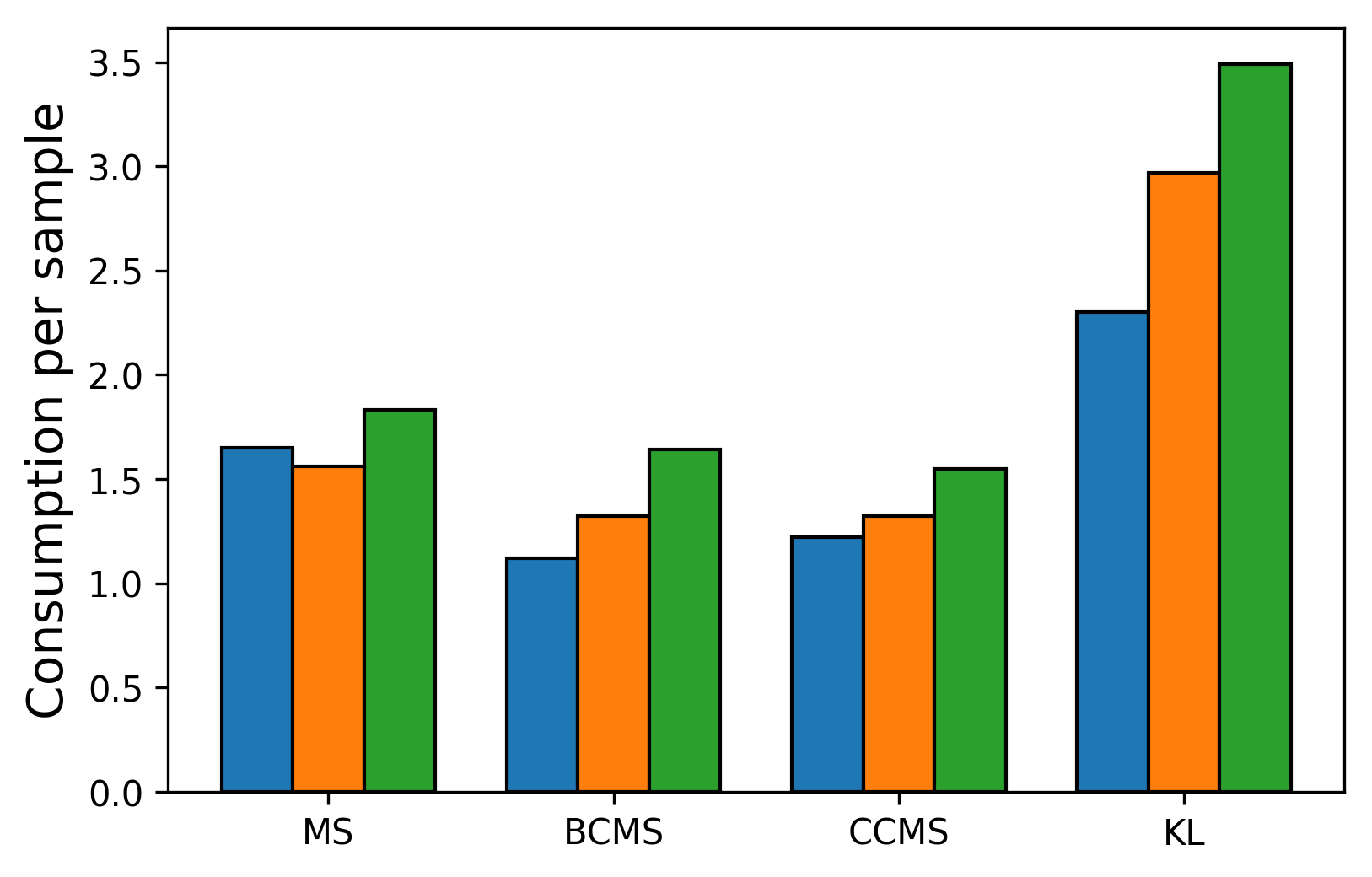}} \\
  \subfloat[Budget constraint residual]{%
      \includegraphics[width=0.45\linewidth]{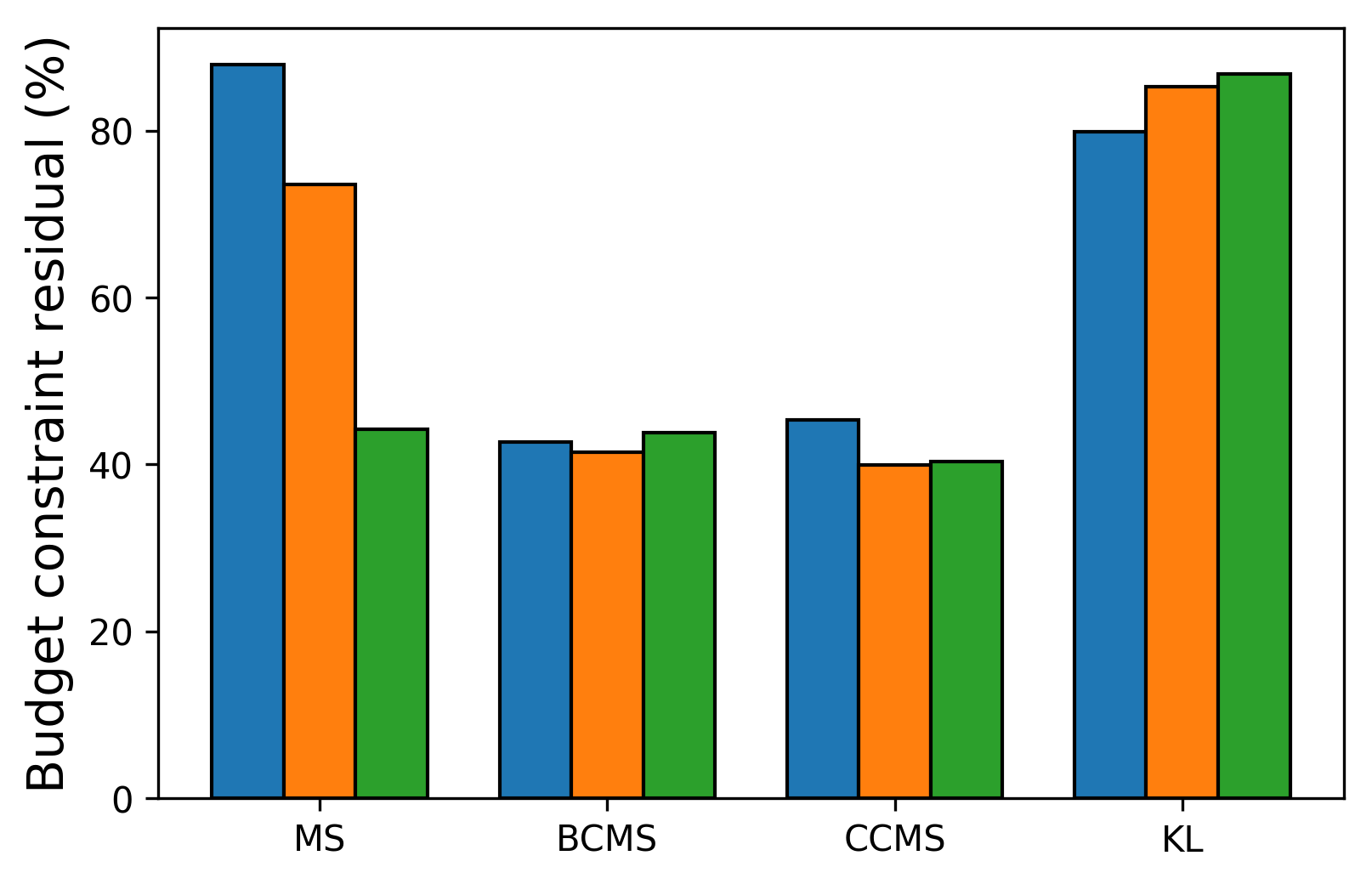}} \hspace{0.5cm}
  \subfloat[Prediction gap average]{%
      \includegraphics[width=0.46\linewidth]{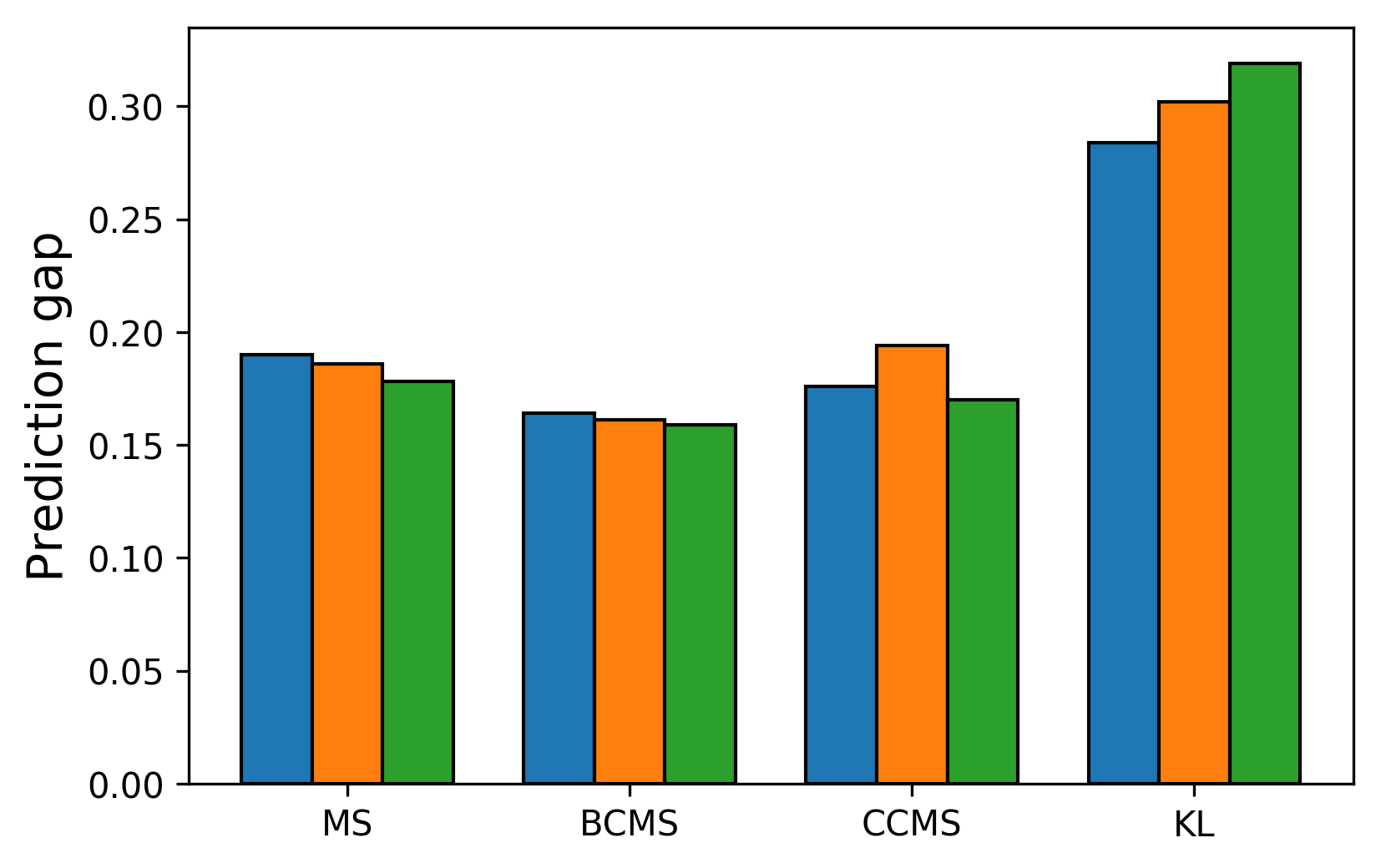}} 
  \caption{Real-world data: Budget experiment} \label{exp:as_bud}
\end{figure}

Figure \ref{exp:as_bud} shows the results of the budget experiment. We find that algorithms with the Gumbel's method BCMS and CCMS perform better than other algorithms. They achieve a larger size of successfully perturbed samples than other algorithms, and they also achieve lower consumption per sample defined as the budget used by all of the samples divided by $|\tilde{S}|$. The relative improvement of BCMS and CCMS over KL is 10\% and 7\%, 9\% and 6\%, and 7\% and 4\% for small, middle, and large budget, respectively. This is because the objective of the max samples models is to maximize the number of successfully perturbed samples. In addition, in plot (a) we observe that a larger budget achieves a larger size of successfully perturbed samples for all algorithms, which is expected. We also analyze budget and prediction confidence constraints. For budget constraint residuals defined by $-g_i$ in \eqref{model:budget} divided by the total available budget, we compute how much of the budget is spent for each budget constraint, and calculate the mean of them, see plot (b). In addition, a prediction gap is computed by measuring the gap between the top and the second best prediction probabilities, see plot (d) for average prediction gaps. We find that budget constraint residuals of BCMS and CCMS are lower than those from other algorithms, and their prediction gaps are smaller than the others. We reason this as BCMS and CCMS spend budgets large enough to guarantee a certain level of prediction confidence; it is larger than $\delta$ in their predictions, but not more than necessary. On the other hand, KL has a large prediction gap that shows high confidence in its prediction which is more than necessary. This is a reason their budget residual per sample is relatively large.

\subsubsection{Scalability experiments}

We also conduct a scalability analysis of our algorithms. We use three different sizes of samples $|{\cal S}| = 300, 600$, and $900$, with samples in each set being grouped into $R=15$ based on their input sequence length.  
In addition, they have inclusive relationships such that ${\cal S}_{300} \subset {\cal S}_{600} \subset {\cal S}_{900}$. In this context, we have two strategies of initializing samples to be perturbed. First, we initialize input features of samples in the larger set with previously obtained values from the subset, and the rest of samples that are not in the subset are initialized randomly. For example, we run an algorithm on ${\cal S}_{300}$, and then run the algorithm on ${\cal S}_{600}$. When we run it on ${\cal S}_{600}$, we initialize samples from ${\cal S}_{300}$ in ${\cal S}_{600}$ with obtained values from the run on ${\cal S}_{300}$, and samples from ${\cal S}_{600} \setminus {\cal S}_{300}$ randomly. The other strategy is to initialize all samples randomly. Similar to the budget experiments, all samples are originally labeled as one of negative classes and correctly predicted by the trained classifier. We use the middle size budget and the other hyperparameters are  the same as those used in the budget experiments. 

\begin{figure} 
    \centering
  \subfloat[Size of $\Tilde{\cal S}$]{%
      \includegraphics[width=0.45\linewidth]{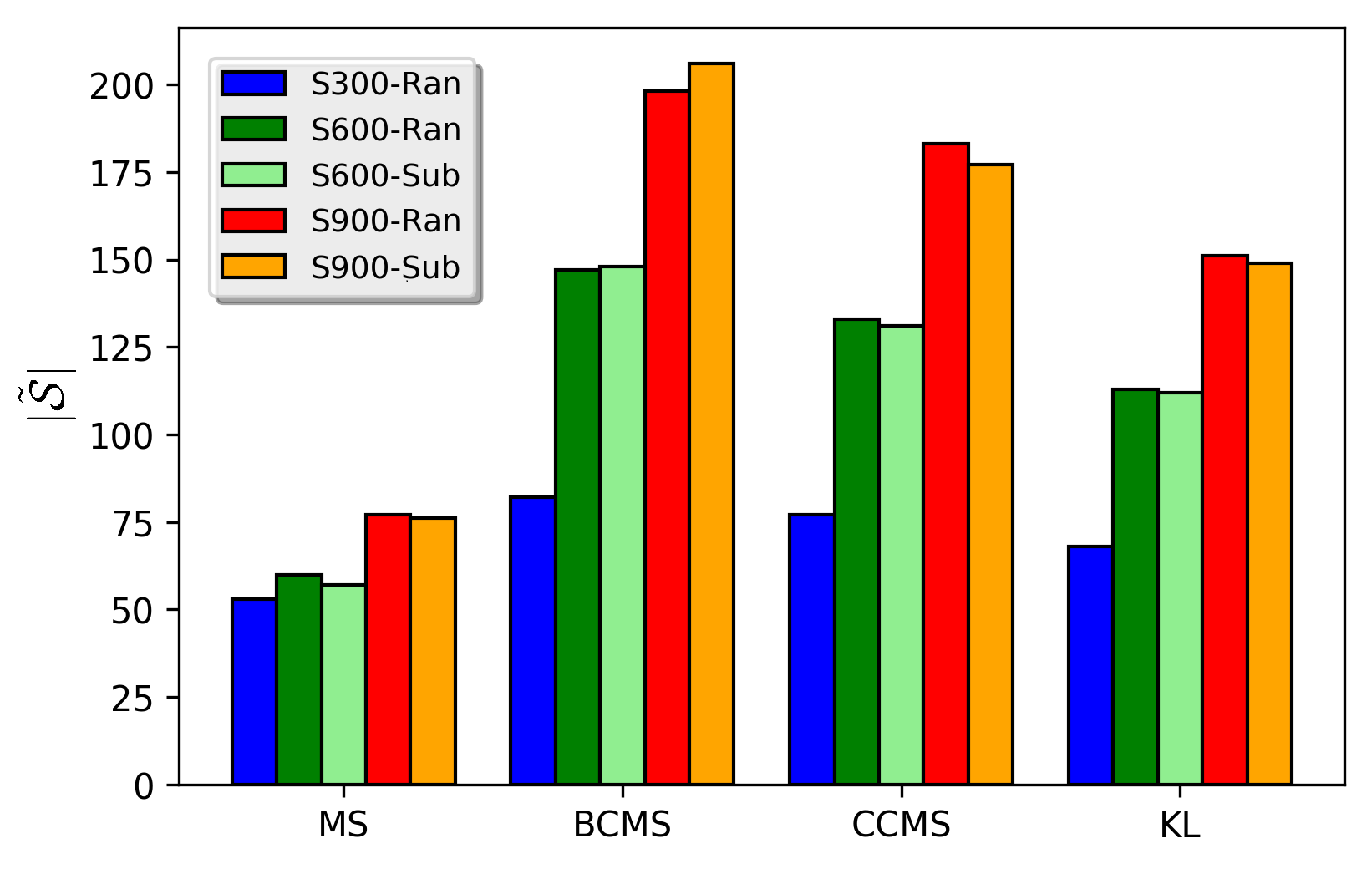}} \hspace{0.5cm}
  \subfloat[Consumption per sample average]{%
      \includegraphics[width=0.44\linewidth]{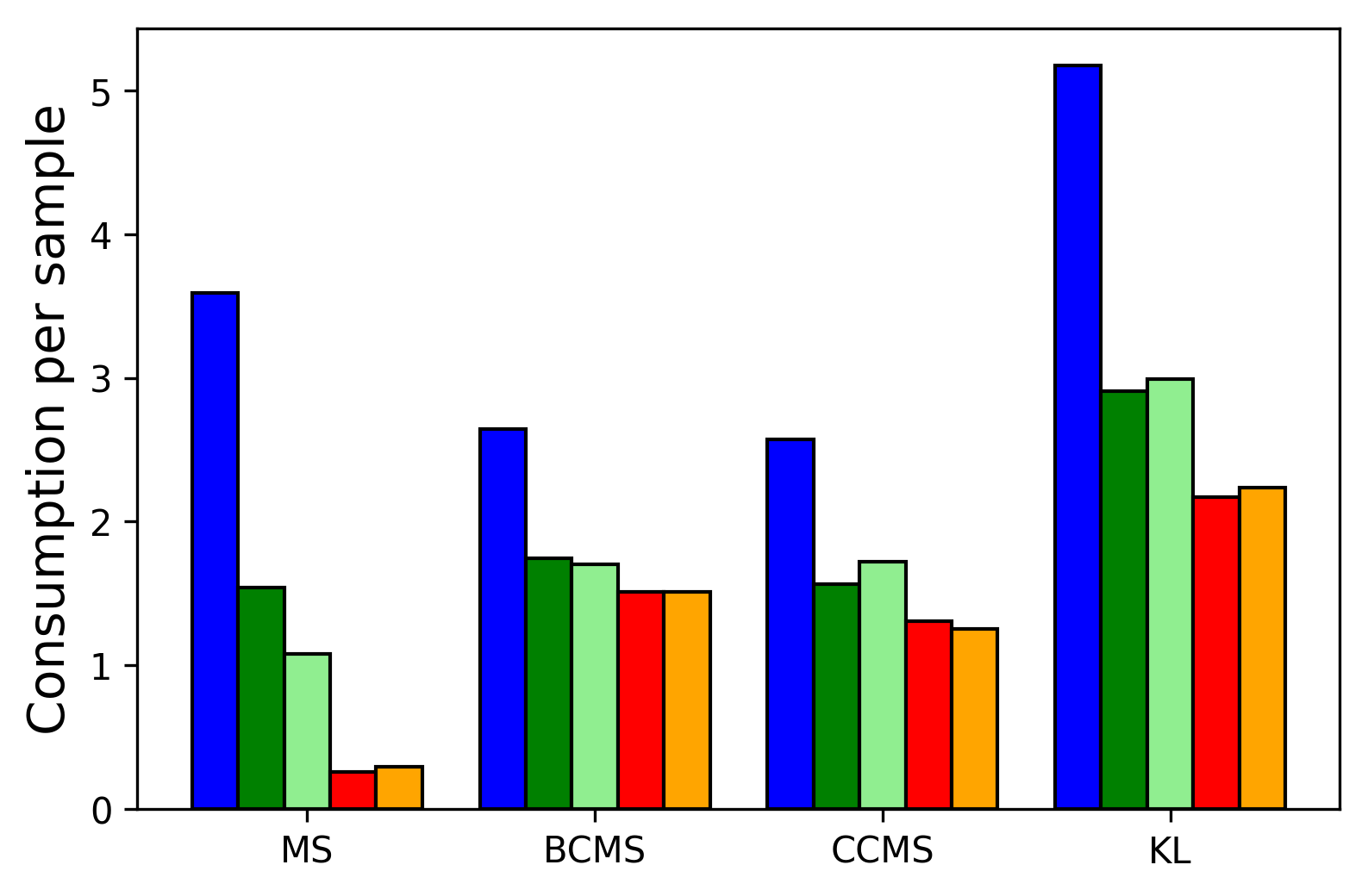}} \\
  \subfloat[Budget constraint residual]{%
      \includegraphics[width=0.45\linewidth]{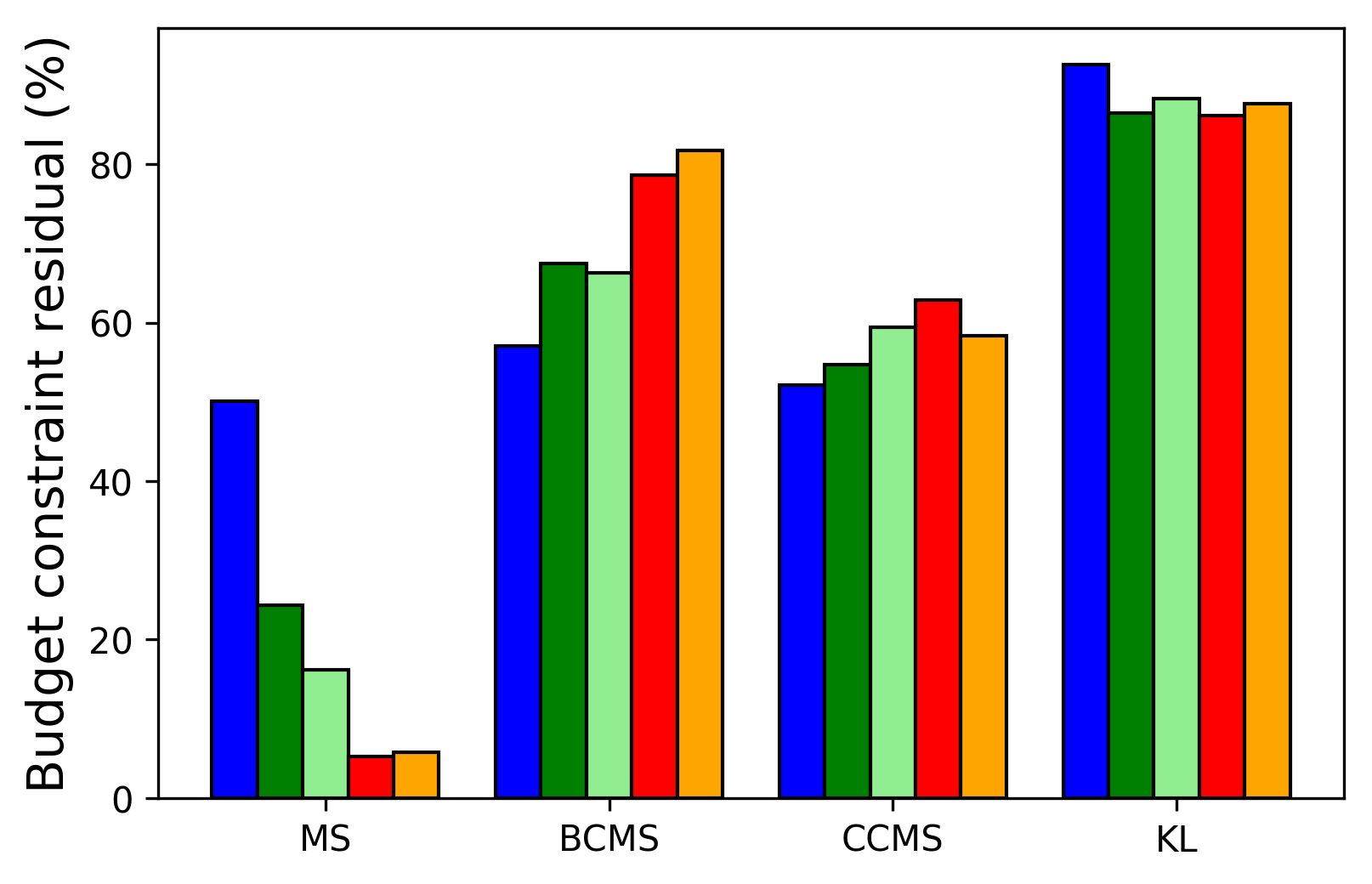}} \hspace{0.5cm}
  \subfloat[Prediction gap average]{%
      \includegraphics[width=0.46\linewidth]{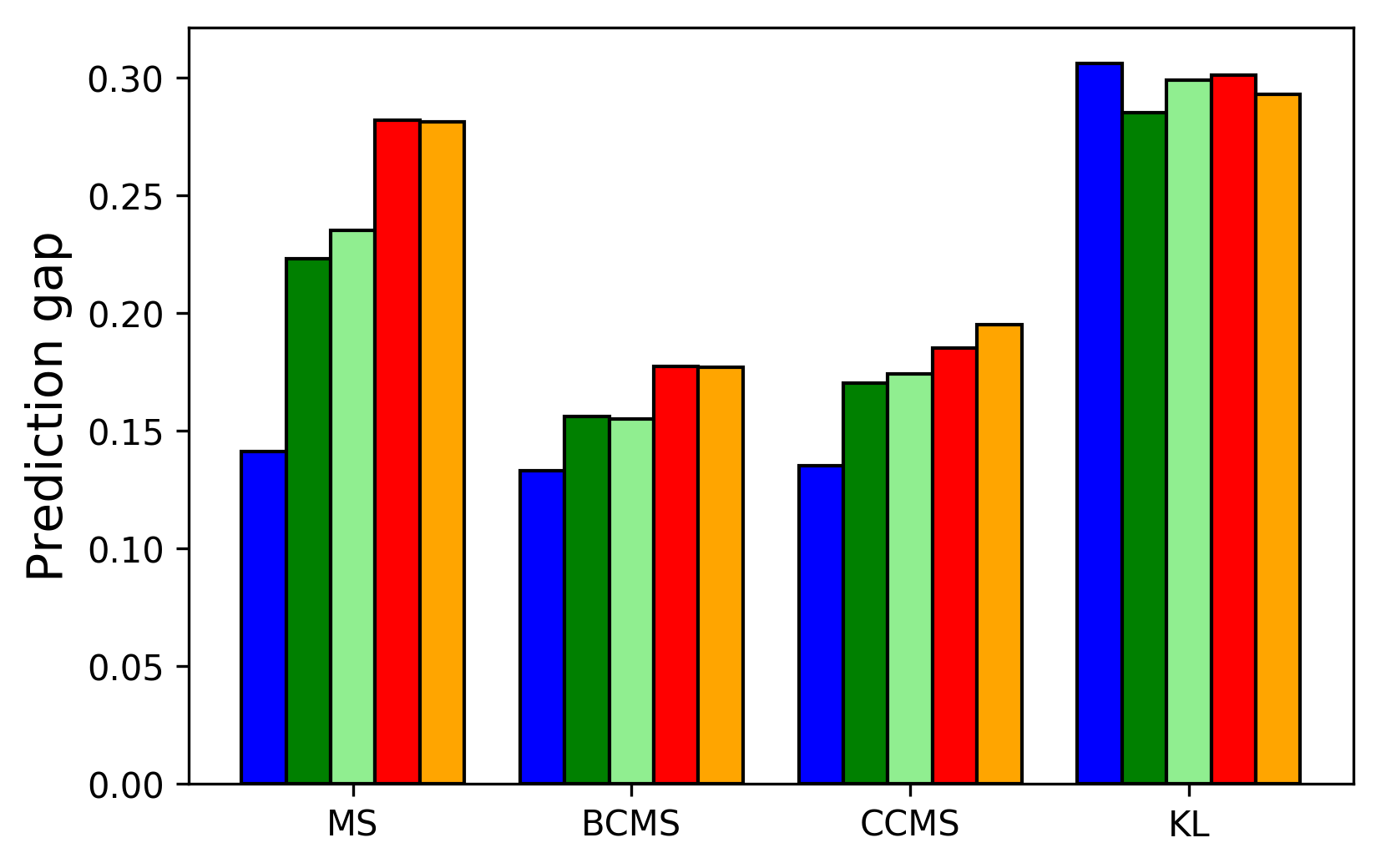}} 
  \caption{Real-world data: Scalability experiment} \label{exp:as_scale}
\end{figure}

\begin{figure} 
    \centering
      \includegraphics[width=0.45\linewidth]{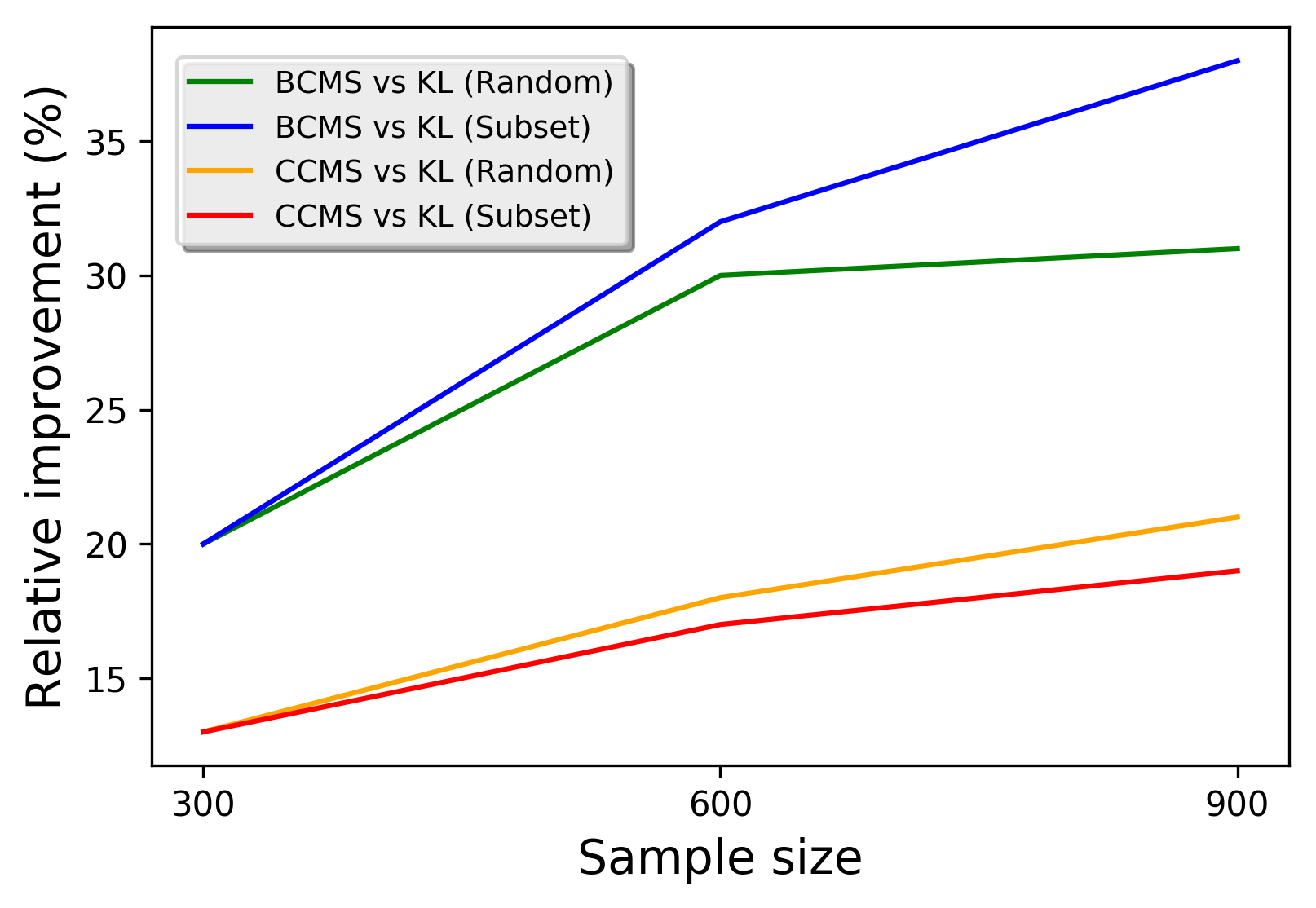}
  \caption{Real-world data: Relative improvement as the sample size increases} \label{exp:as_scale_improvement}
\end{figure}

Figure \ref{exp:as_scale} shows the results of the scalability experiment. Note that a run on ${\cal S}_{300}$ with subset initialization is denoted by $\text{300-Sub}$, and one with random initialization is denoted by $\text{300-Ran}$ in the figure. Similar to the budget experiments, the algorithms with Gumbel's method BCMS and CCMS perform better than other algorithms. The count of samples successfully perturbed  by them is larger than the counts from other algorithms, and also the average consumption per sample is smaller.  The relative improvement of the stochastic models BCMS and CCMS over KL is 20\% and 13\% for $|{\cal S}| = 300$, 31\% and 17.5\% for $|{\cal S}| = 600$, and 34.5\% and 20\% for $|{\cal S}| = 900$ on average over different initialization strategies. The observations from Section \ref{sec:as_bud} apply to each individual sample size. We find that the relative improvement of BCMS and CCMS over KL increases linearly as sample size increases in Figure \ref{exp:as_scale_improvement}. In terms of the two initialization strategies, both cases show similar results and thus the benefits of warm-start are negligible. Regarding the budget and prediction confidence constraints, we find similar results to the budget experiments. Budget constraint residuals for BCMS and CCMS are lower than KL, and their prediction gaps are smaller than for the other algorithm. The aforementioned conclusions apply to all of different sizes of samples and thus we conclude that our algorithms scale efficiently. 

\subsection{Public data: MIMIC}

MIMIC is a public dataset that describes clinical information of patients admitted to an Intensive Care Unit at the Beth Israel Deaconess Medical Center in Boston, Massachusetts from 2001 to 2012. It contains 58,576 samples for patient admissions. Descriptive statistics can be found in \cite{mimic2,mimic1}. In this study, we use 13 input features corresponding to the health state for 30-day mortality predictions used in \cite{Luo2018}. Since MIMIC is a time series dataset and has missing values, we use a recurrent network based on Gated Recurrent Unit, called GRU-D, that is widely used for coping with multivariate time series with missing values \cite{Che2018} for imputation and predictions. Its AUC is around 0.78 which is comparable to state-of-the-art. We conduct only a budget experiment for this dataset due to its limited size. 

We perturb 75 samples selected from the test set, i.e., $|{\cal S}| = 75$. The 75 samples are originally labeled as ``dead'' and correctly predicted by the trained classifier. Our purpose is to perturb the samples so that they are predicted as ``alive.'' To decide the size of budgets which represent the per drug amount, we first run  Algorithm \ref{alg:KL} with unlimited budget constraints to measure how much perturbation is needed. Then, we determine small, middle, and large sizes of budgets by imposing 40\%, 60\%, and 80\% of the total budget achieved by unlimited budgets. 

\begin{figure}
    \centering
  \subfloat[Size of $\Tilde{\cal S}$]{%
      \includegraphics[width=0.45\linewidth]{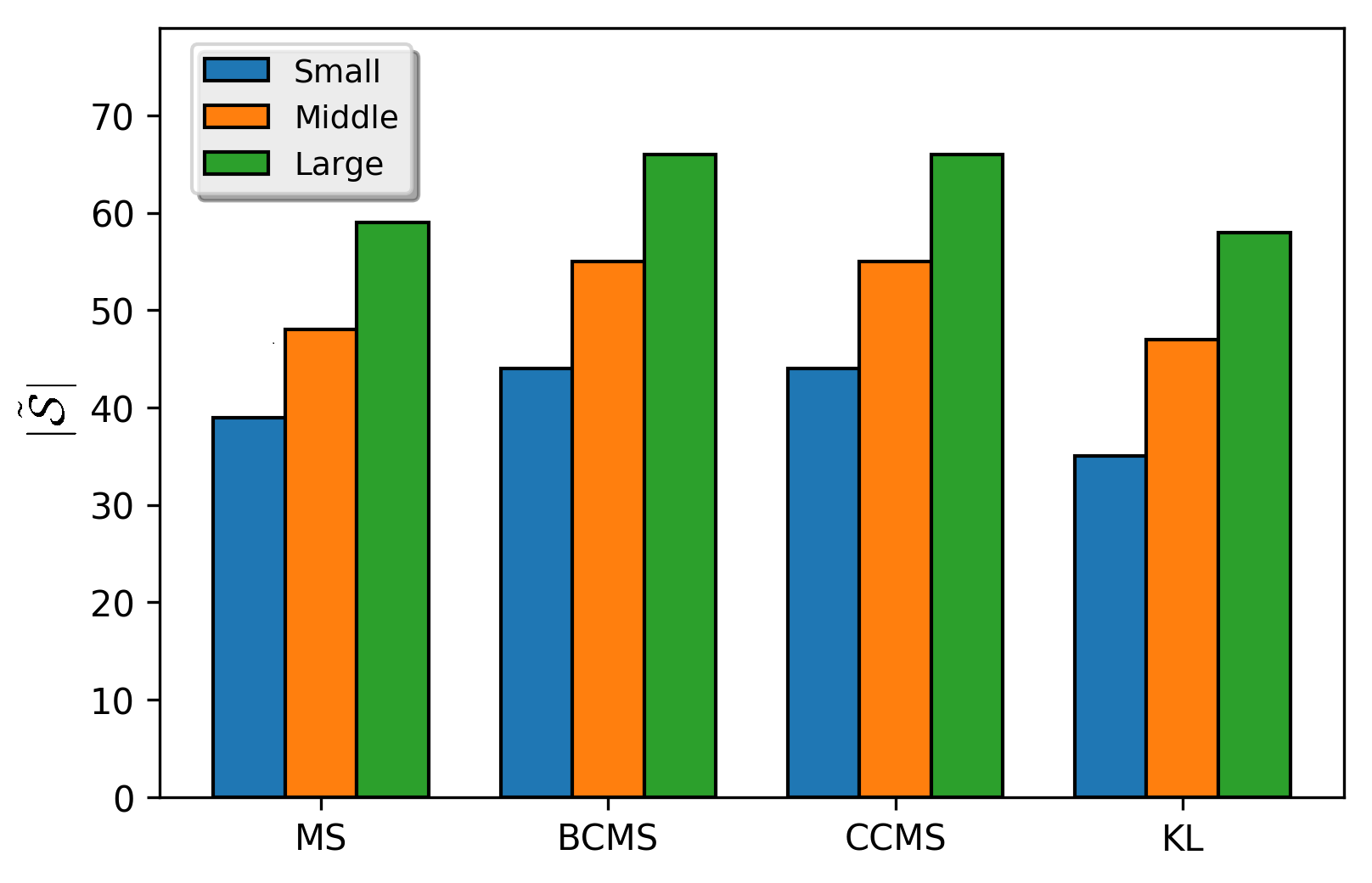}} \hspace{0.5cm}
  \subfloat[Consumption per sample average]{%
      \includegraphics[width=0.45\linewidth]{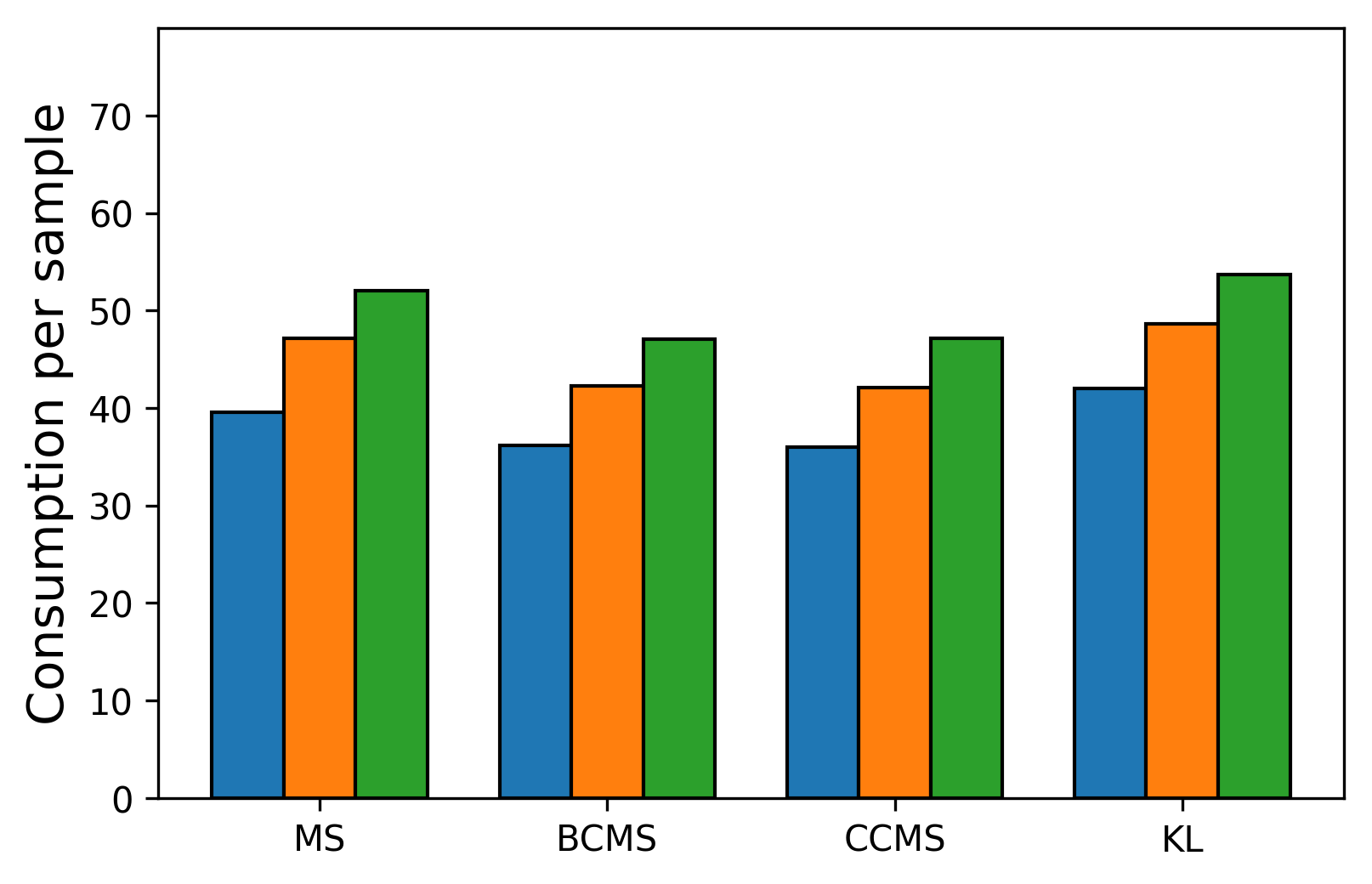}} \\
  \subfloat[Budget constraint residual]{%
      \includegraphics[width=0.45\linewidth]{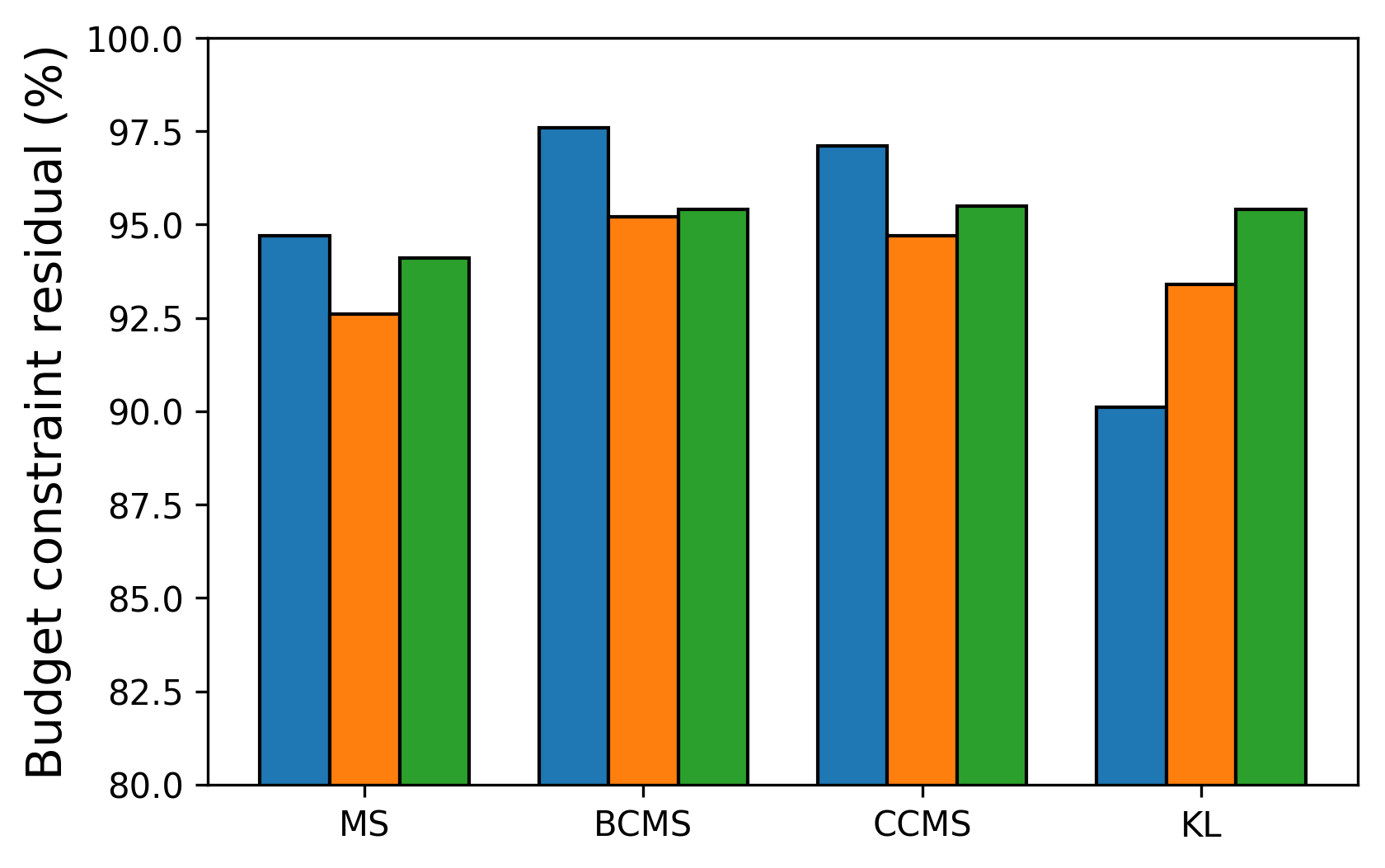}} \hspace{0.5cm}
  \subfloat[Prediction gap average]{%
      \includegraphics[width=0.45\linewidth]{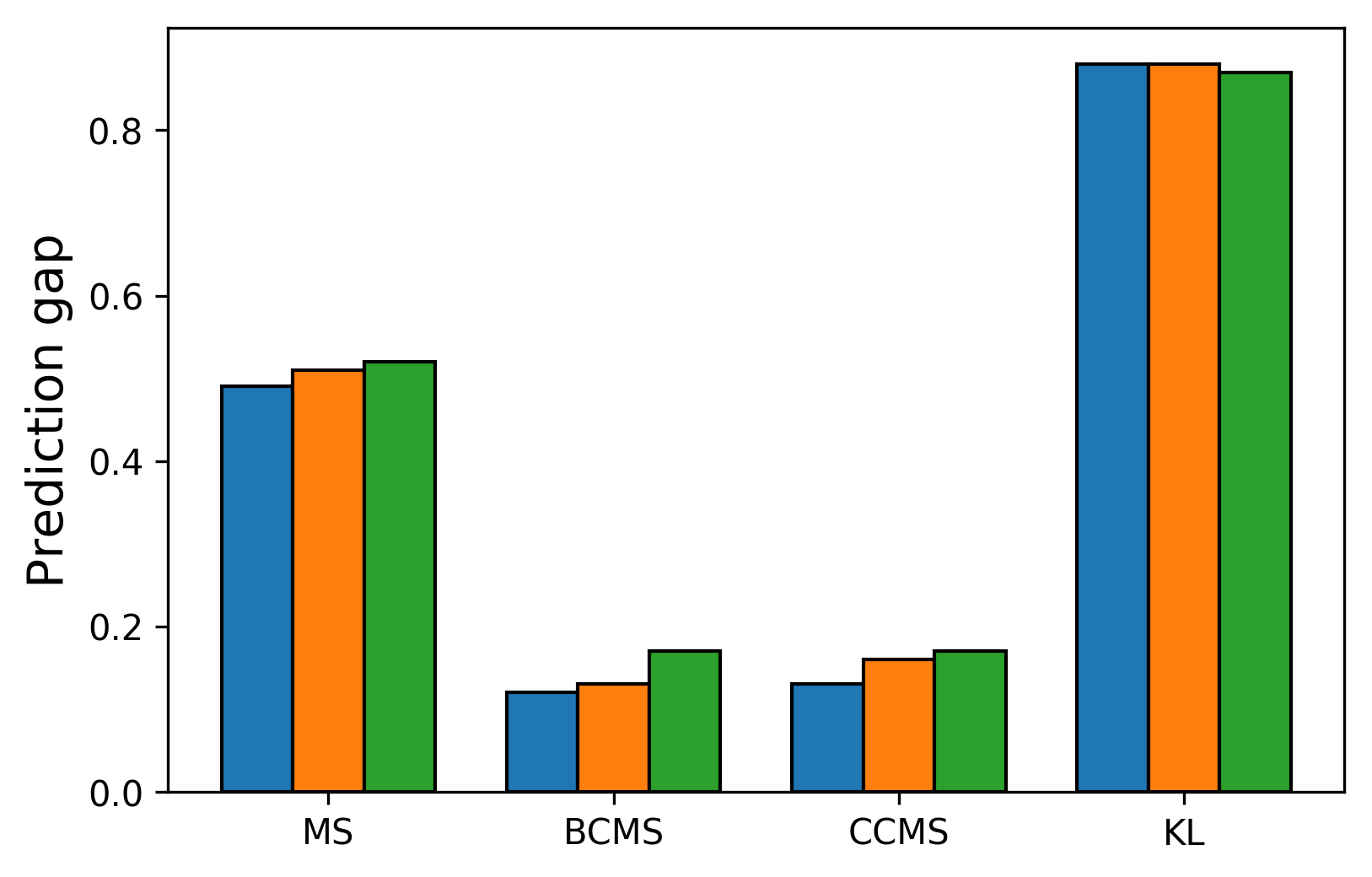}} 
  \caption{MIMIC: Budget experiment} \label{exp:mimic_bud} 
\end{figure}

Figure \ref{exp:mimic_bud} shows the results of the budget experiment. Similar to the results on the real-world data, stochastic algorithms for max samples models, BCMS and CCMS perform better than KL and MS. They obtain a larger size of successfully perturbed samples than the other algorithms, and also achieve smaller consumption per sample. The relative improvement of our max samples models MS, BCMS and CCMS over KL is 5\%, 19\% and 19\% on average for all different budget scenarios. It is interesting to observe that KL uses a much bigger portion of the budget than the other algorithms.  

\subsection{Optimization aspects}

\begin{figure}
    \centering
  \subfloat[MIMIC: Budget experiment]{%
      \includegraphics[width=0.77\linewidth]{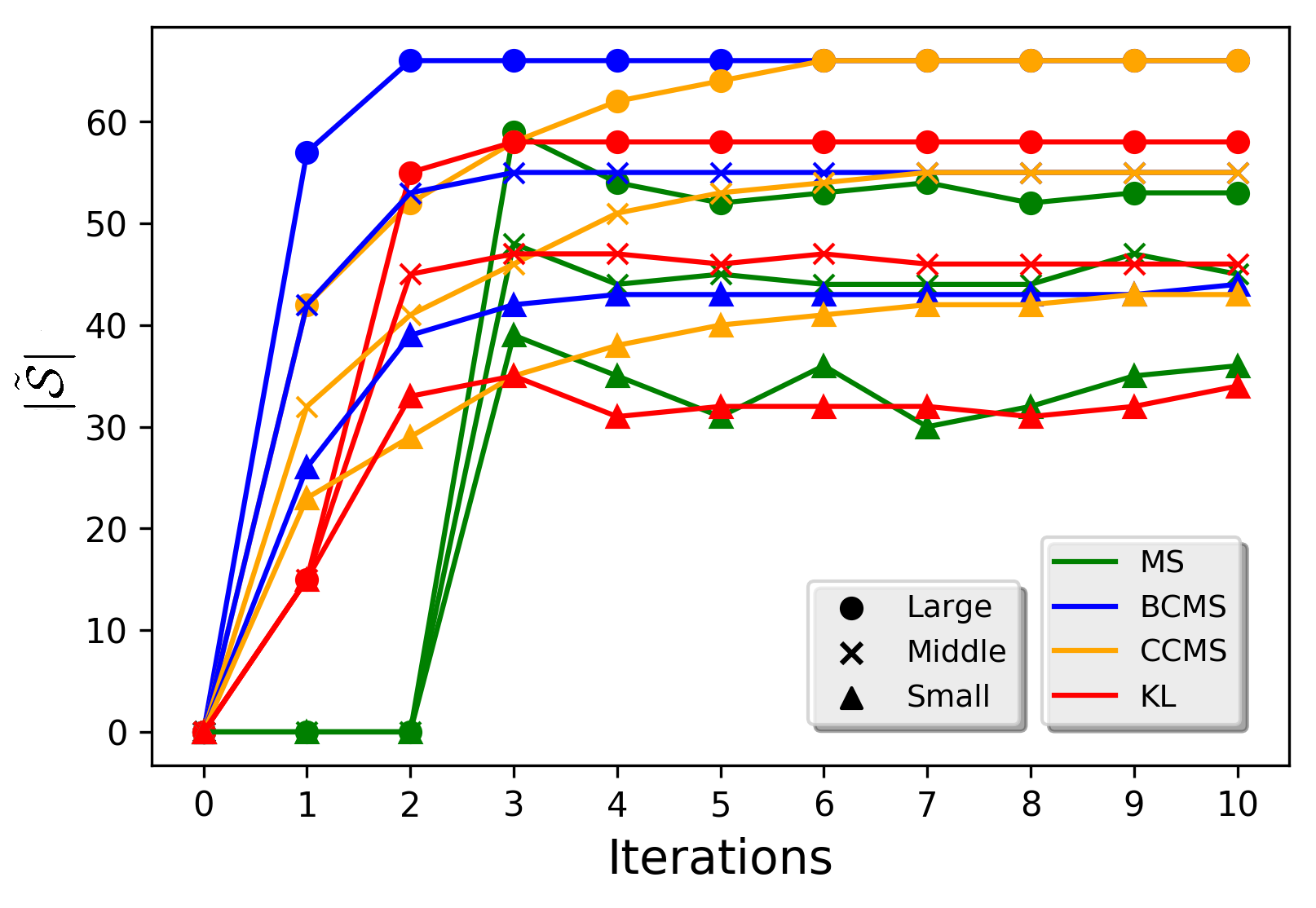}} \\
  \subfloat[Real-world data: Budget experiment]{%
      \includegraphics[width=0.76\linewidth]{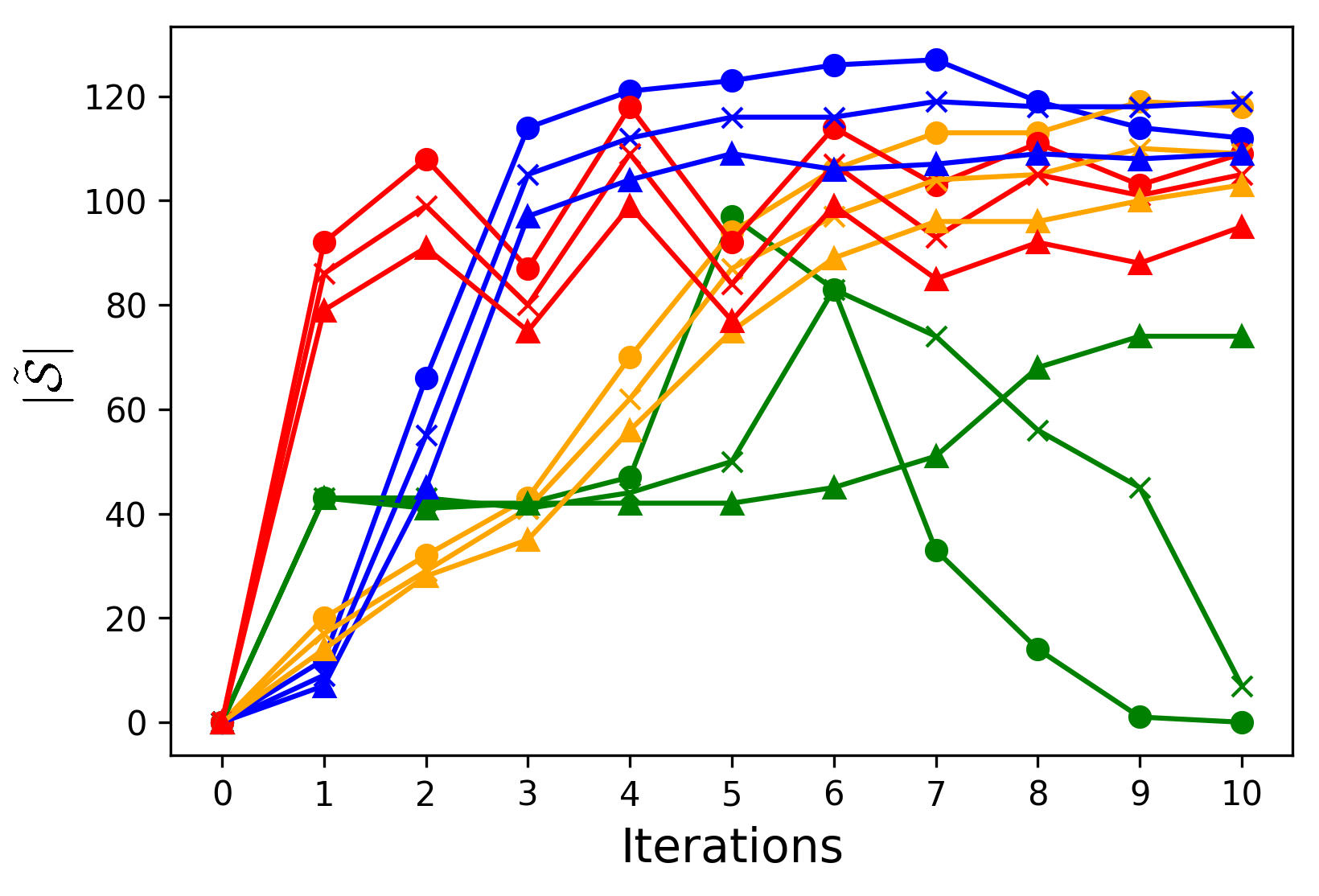}} \\
  \subfloat[Real-world data: Scalability experiment]{%
      \includegraphics[width=0.76\linewidth]{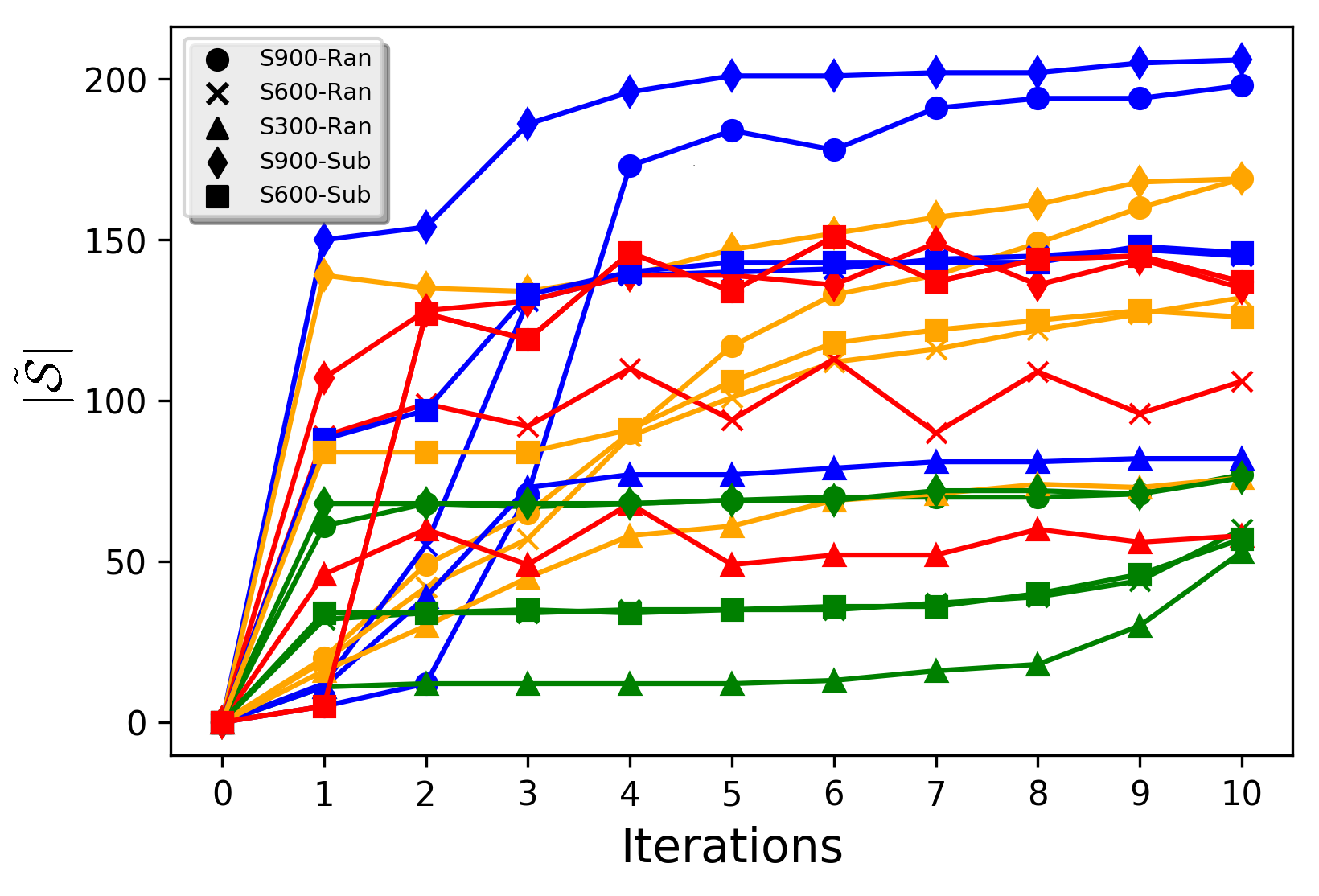}} 
  \caption{Size of $\Tilde{\cal S}$ at each outer iteration} \label{exp:stilde_opt} 
\end{figure}

In this part we discuss optimization aspects of the algorithms. Figure \ref{exp:stilde_opt} shows sizes of $\Tilde{\cal S}$ (successfully perturbed samples) for each algorithm at each outer iteration. We observe that the number of $\Tilde{\cal S}$ increases with iterations and stays at the highest point, which implies that the algorithms converge to a (local) optimal value. We also find that BCMS converges faster than CCMS in all cases. We also note that KL converges faster than CCMS, however, the largest $|\Tilde{{\cal S}}|$ from KL is not larger than the one from CCMS and BCMS. 

\begin{figure}
    \centering
  \subfloat[$||\nabla_{\mu} \, L||$ of MS]{%
      \includegraphics[width=0.8\linewidth]{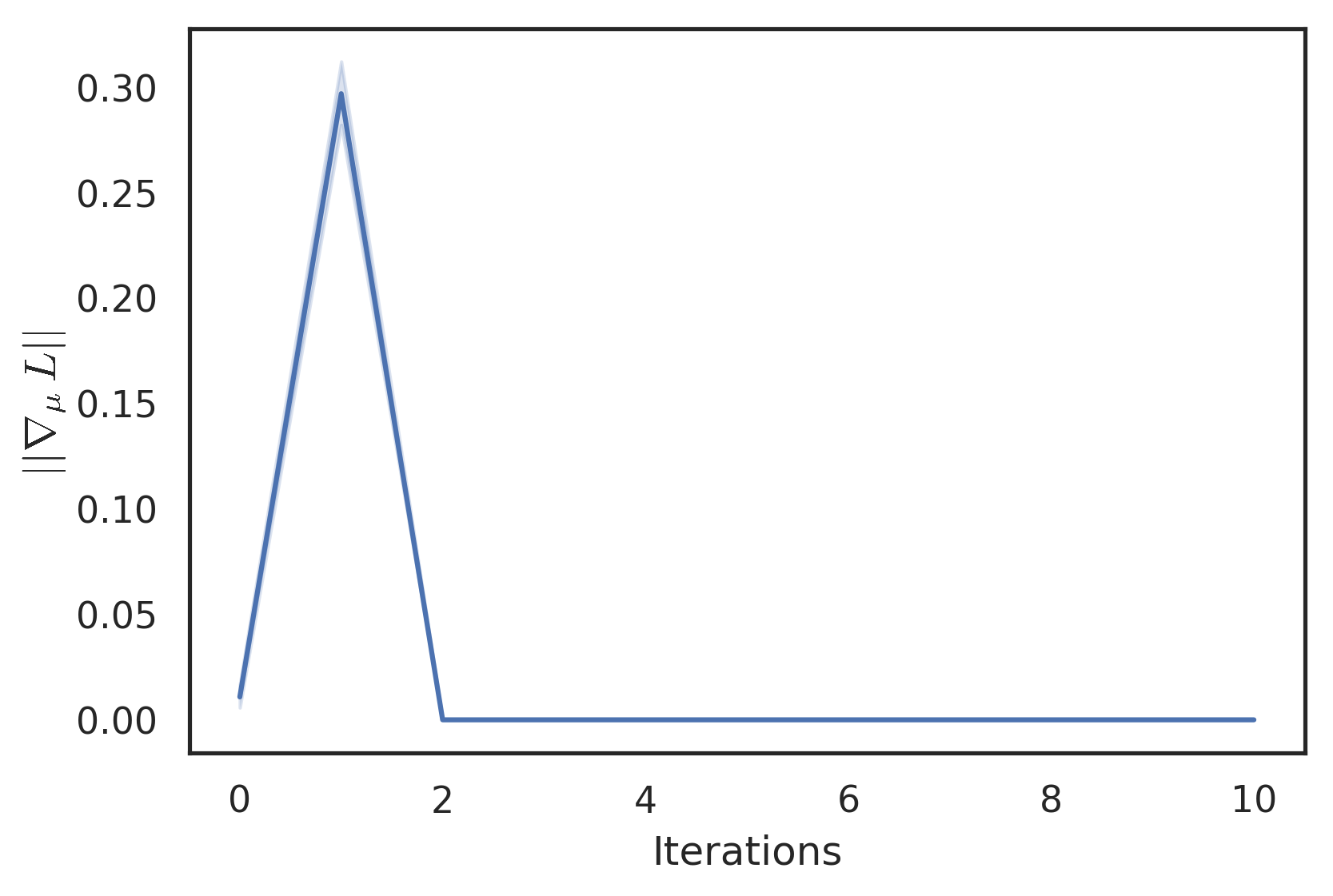}} \\
  \subfloat[$|| \nabla_{\mu} \, L||$ of BCMS]{%
      \includegraphics[width=0.8\linewidth]{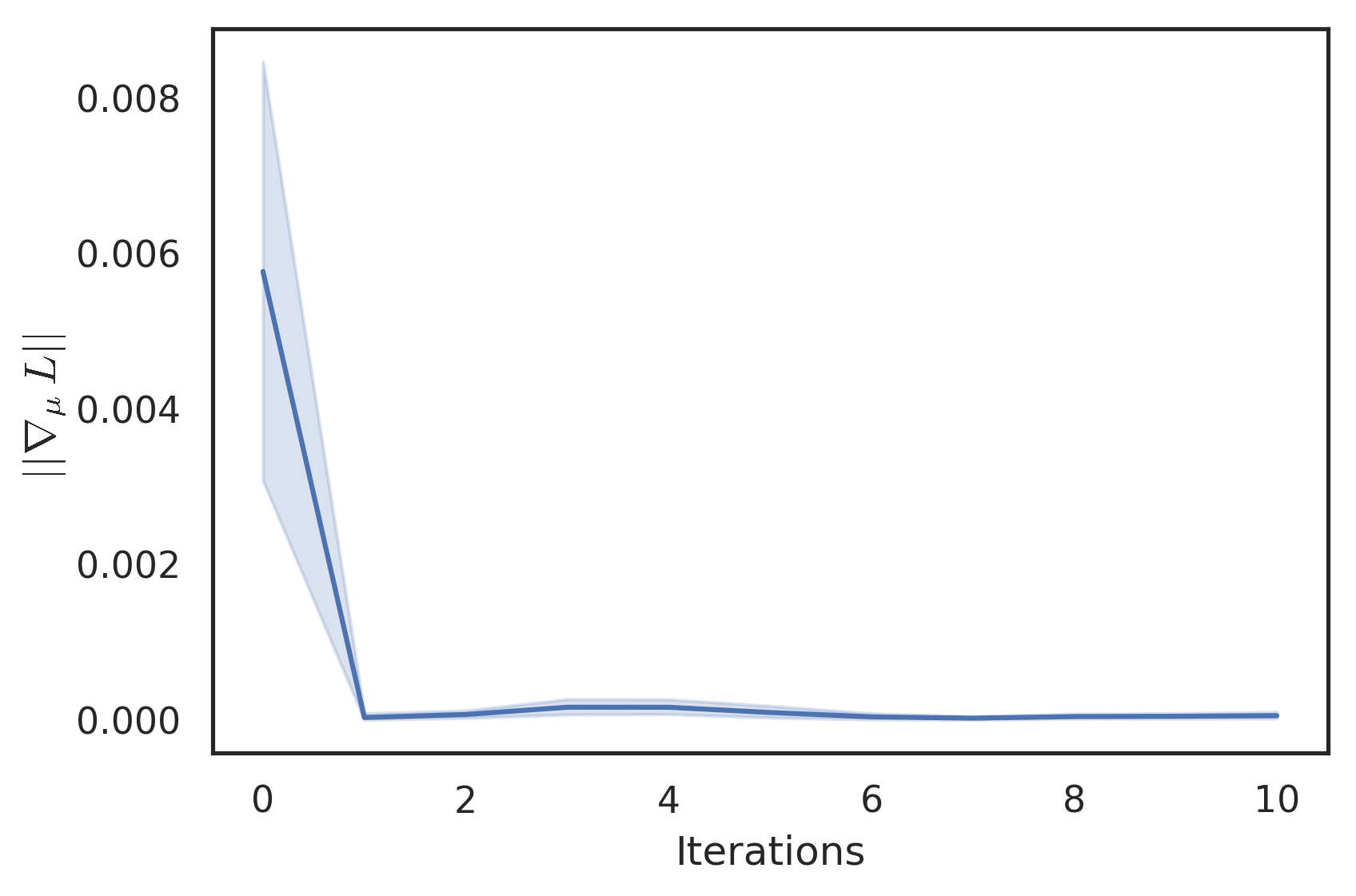}} \\
  \subfloat[$|| \nabla_{\mu} \, L||$ of CCMS]{%
      \includegraphics[width=0.8\linewidth]{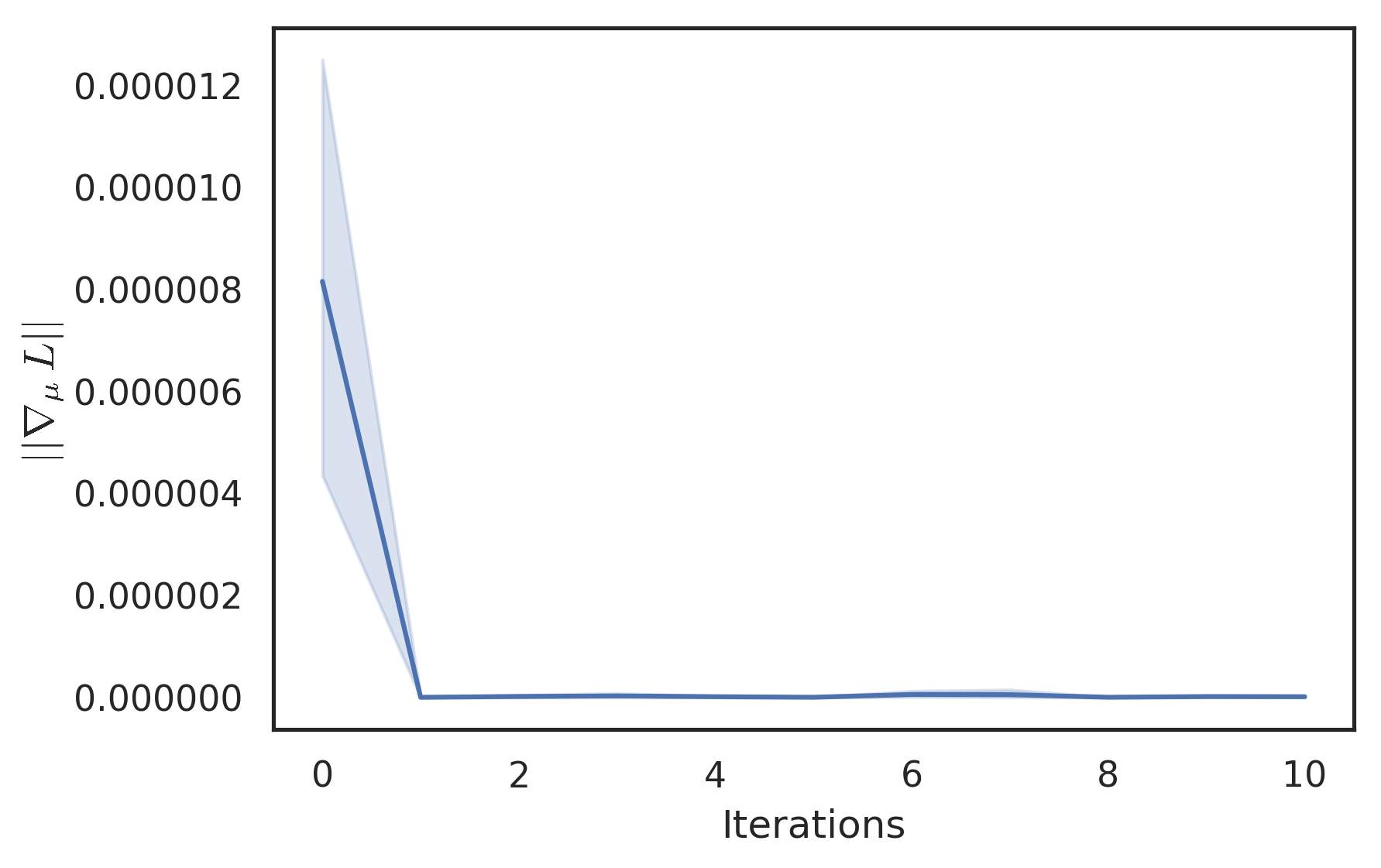}} 
  \caption{Norm of Lagrangian multipliers of max samples algorithms at each outer iteration of MIMIC} \label{exp:lag_grad}
\end{figure}

\begin{figure}
    \centering
  \subfloat[$\lambda$]{%
      \includegraphics[width=0.45\linewidth]{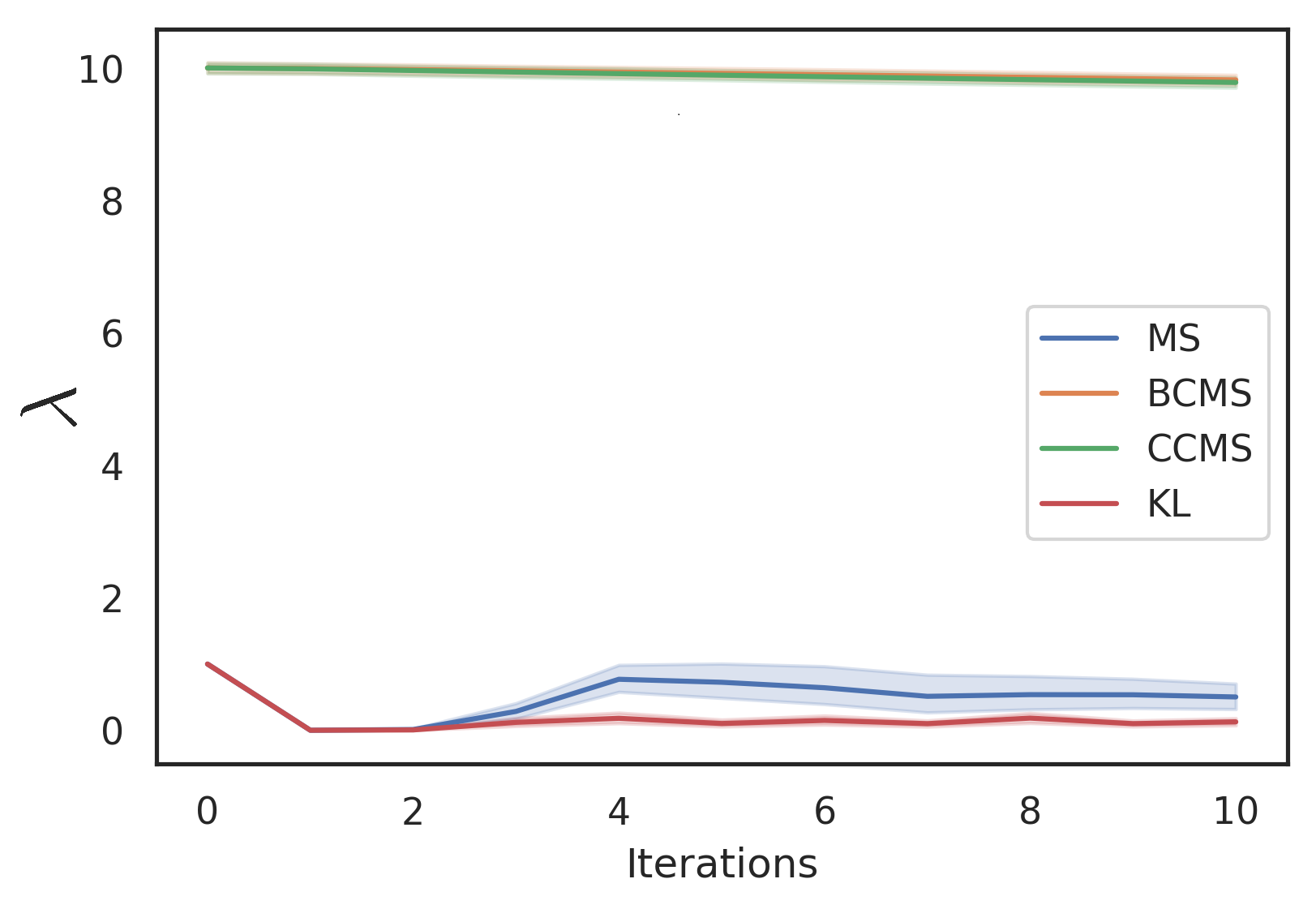}} \vspace{0.5cm} 
  \subfloat[$\mu$]{%
      \includegraphics[width=0.45\linewidth]{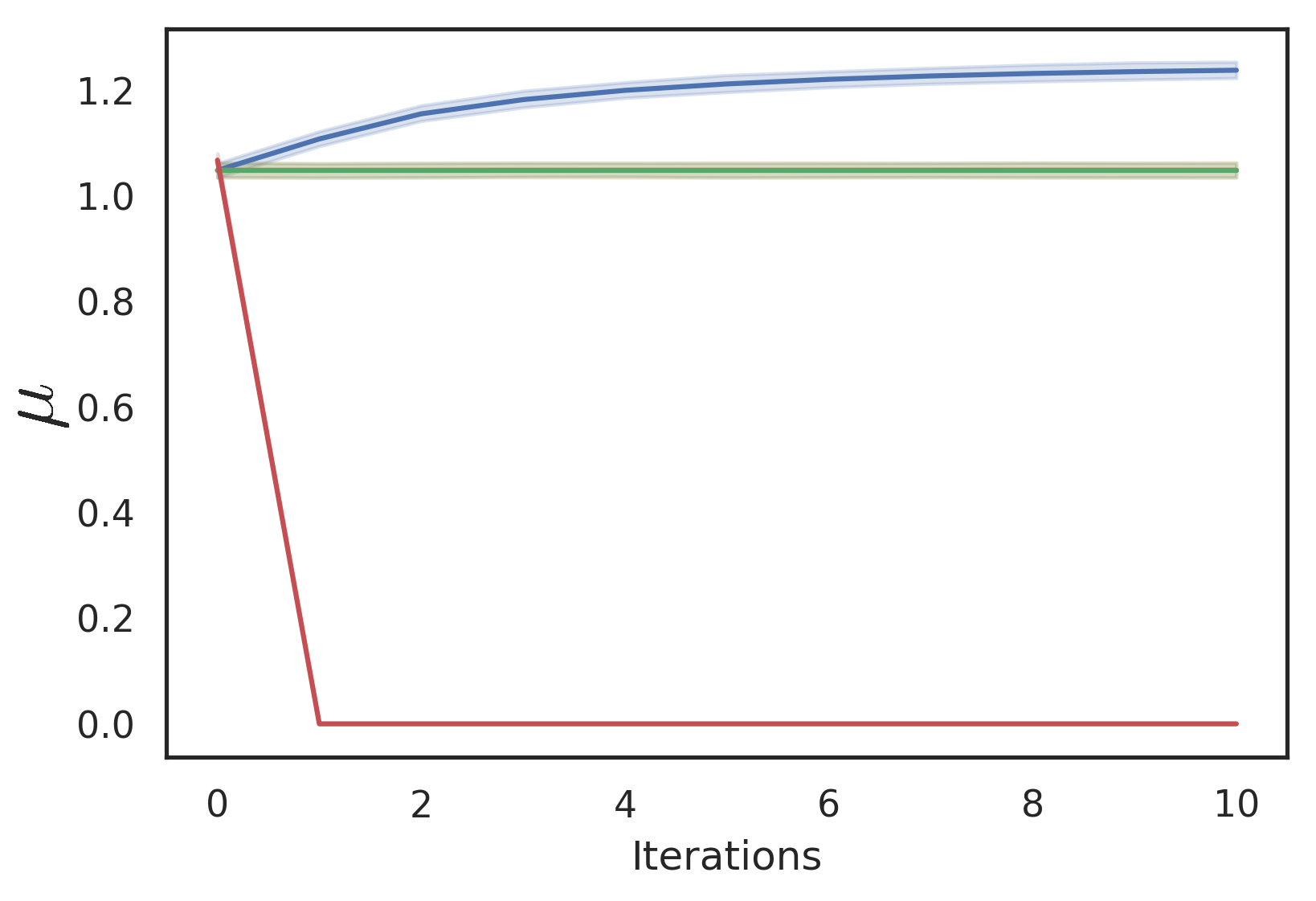}} \hspace{0.5cm} 
  \caption{Lagrangian multipliers at each outer iteration of MIMIC} \label{exp:lag} 
\end{figure}

Figure \ref{exp:lag} shows the values of Lagrangian multipliers at each outer iteration for the MIMIC dataset. The average values of $\lambda$ with respect to the budget constraint decrease as each algorithm iterates, which implies that the amount of constraint violation decreases. Regarding $\mu$ that is associated with the prediction confidence constraint, KL values decrease as the algorithm proceeds while the max sample algorithms drive them higher. 
In Figure \ref{exp:lag_grad}
we further observe that the norm of gradients of $\mu$ for the max samples algorithms shrink to zero, which implies that the algorithms reduce the amount of constraint violation. 
MS shows erratic behaviors in some cases and its performance is unstable. We reason this is due to the presence of constraints and binary variables in MS. 

Table \ref{exp:runtime} shows the actual runtime of the algorithms on the larger, proprietary dataset. Algorithms for stochastic models BCMS and CCMS take longer to update variables per iteration in the inner loop than MS and KL require since the Gumbel's simulation is implemented. However, the total runtime of BCMS and CCMS does not necessarily take longer than others. Based on the runtime, convergence and size of $\Tilde{\cal S}$ we conclude that BCMS is the best performer among all the algorithms considered herein. Algorithms CCMS is a close second. 

\begin{table}[H]
\begin{center}
\caption{Runtime of algorithms} \label{exp:runtime}
\begin{tabular}{cccc}
\hline
\multirow{2}{*}{Algorithm} & \multicolumn{2}{c}{Runtime per iteration (No. of the max iterations)} & \\ 
     & Outer loop & Inner loop & Total runtime \\
\hline
MS   & 20 sec (10) & 0.7 sec (10,000) & 19 hr \\ 
BCMS & 25 sec (10) & 18 sec (100)     & 5 hr \\ 
CCMS & 25 sec (20) & 15 sec (100)     & 8 hr \\ 
KL   & 20 sec (10) & 0.9 sec (5,000)  & 10 hr \\ 
\hline
\end{tabular}
\end{center}
\end{table}

\section{Conclusion} \label{sec:conclusion}
In this paper, a new framework for inverse classification is proposed. We formulate a constrained optimization problem that maximizes the number of successfully perturbed samples with budget and prediction confidence constraints. In addition, we formulate a stochastic problem with chance constraints. To solve the constrained problems, algorithms based on Lagrangian and subgradient methods are developed. Based on the computational study, we find that the algorithms perform well in various budget settings and are scalable. 

\twocolumn
\onecolumn

\bibliographystyle{abbrv}  
\bibliography{main}  

\appendix

\end{document}